\documentclass[journal,compsoc]{IEEEtran}

\usepackage{soul,framed} 
\usepackage[table]{xcolor}
\colorlet{shadecolor}{yellow}
\usepackage[pdftex]{graphicx}
\graphicspath{{../pdf/}{../jpeg/}}
\DeclareGraphicsExtensions{.pdf,.jpeg,.png}

\usepackage{marvosym}
\usepackage[cmex10]{amsmath}
\usepackage{array}
\usepackage{mdwmath}
\usepackage{mdwtab}
\usepackage{eqparbox}
\usepackage{url}
\usepackage{graphicx}
\usepackage{amsmath}
\usepackage{amsfonts}
\usepackage{algorithm}
\usepackage{algpseudocode}
\usepackage{multirow}
\usepackage{footnote}
\usepackage{bm}
\usepackage{makecell}
\usepackage{pdflscape}
\usepackage{afterpage}
\usepackage{lscape}
\usepackage{booktabs}
\usepackage{tikz}

\usepackage{pgfplots}
\pgfplotsset{compat=1.12}
\usepackage{subcaption}
  \pgfplotsset{
    layers/my layer set/.define layer set={
      background,
      main,
      foreground
    }{ },
    set layers=my layer set,
  }

\usepackage[square,sort,comma,numbers,sort&compress]{natbib} 
\usepackage[top=0.8in,bottom=0.8in,left=0.5in,textwidth=7.5in]{geometry}

\usepackage[pagebackref=false,breaklinks=false,linkcolor=red,anchorcolor=black, citecolor=black,colorlinks,bookmarks=true]{hyperref}

\usepackage{amsfonts}

\usepackage{pifont}

\definecolor{mygray}{gray}{.55}
\definecolor{mypink}{RGB}{244,0,122}
\definecolor{mygreen}{RGB}{0,104,0}
\definecolor{myyellow}{RGB}{255,102,0}
\definecolor{DarkBlue}{rgb}{0,0,1}
\usepackage{caption}
\usepackage{graphicx}
\usepackage{amsmath}
\usepackage{amssymb}
\usepackage{colortbl}

\makeatletter
\def\thanks#1{\protected@xdef\@thanks{\@thanks
		\protect\footnotetext{#1}}}
\makeatother

\begin{document}
	
\title{Visible-Thermal Tiny Object Detection: A Benchmark Dataset and Baselines}

\author{Xinyi~Ying, Chao~Xiao, Wei~An, Ruojing~Li, Xu~He, Boyang~Li, Xu~Cao, Zhaoxu~Li, Yingqian~Wang, Mingyuan~Hu, Qingyu~Xu, Zaiping~Lin, Miao~Li, Shilin~Zhou, Weidong~Sheng, Li~Liu
	\IEEEcompsocitemizethanks{
		\IEEEcompsocthanksitem This work was supported by the National Key Research and Development Program of China No. 2021YFB3100800, the National Natural Science Foundation of China under Grant 62376283, the Science and
		Technology Innovation Program of Hunan Province under Grant 2021RC3069, and the Independent Innovation Science Fund Project of the National University of Defense Technology (NUDT) under Grant 22-ZZCX-042. 
		All authors are with the College of Electronic Science and Technology, NUDT, Changsha, 410073, China. 
		Xinyi~Ying (yingxinyi18@nudt.edu.cn) and Chao~Xiao (xiaochao12@nudt.edu.cn) share the equal contribution. 
		Li~Liu (liuli\_nudt@nudt.edu.cn) and Weidong~Sheng (shengweidong@nudt.edu.cn) are the corresponding authors.}}


\IEEEtitleabstractindextext{
\begin{center}\setcounter{figure}{0}
	\centering
	\vspace{-.5cm}
	\includegraphics[width=0.92\textwidth]{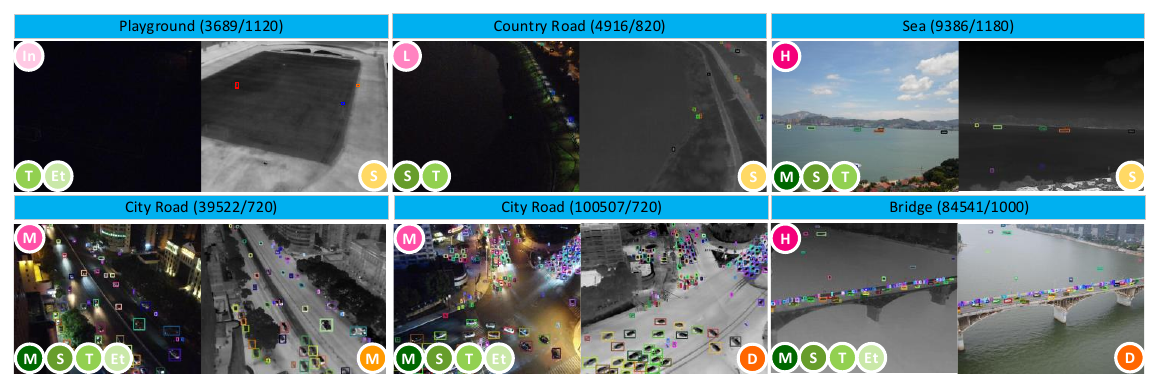}
	\captionof{figure}{
		Example frames of RGBT-Tiny dataset. \textit{Scenes (annotation / frame number)} are shown on the top. Sequence-level attributes are shown at the bottom. \textcolor{mypink}{Pink}, \textcolor{mygreen}{green} and \textcolor{myyellow}{yellow} circles represent levels of light vision (\textit{i.e.,} H: high, M: medium, L: low, In: invisible), target size (\textit{i.e.,} Et: extremely tiny, T: tiny, S: small, M: medium, L: large) and annotation density (\textit{i.e.,} S: sparse, M: medium, D: dense).
	}\label{show}
\vspace{0.1cm}
\end{center}

\begin{abstract}
	Visible-thermal small object detection (RGBT SOD) is a significant yet challenging task with a wide range of applications, including video surveillance, traffic monitoring, search and rescue. However, existing studies mainly focus on either visible or thermal modality, while RGBT SOD is rarely explored. Although some RGBT datasets have been developed, the insufficient quantity, limited diversity, unitary application, misaligned images and large target size cannot provide an impartial benchmark to evaluate RGBT SOD algorithms. In this paper, we build the first large-scale benchmark with high diversity for RGBT SOD (namely RGBT-Tiny), including 115 paired sequences, 93K frames and 1.2M manual annotations. RGBT-Tiny contains abundant objects (7 categories) and high-diversity scenes (8 types that cover different illumination and density variations). Note that, over 81\% of objects are smaller than 16$\times$16, and we provide paired bounding box annotations with tracking ID to offer an extremely challenging benchmark with wide-range applications, such as RGBT image fusion, object detection and tracking. In addition, we propose a scale adaptive fitness (SAFit) measure that exhibits high robustness on both small and large objects. The proposed SAFit can provide reasonable performance evaluation and promote detection performance. Based on the proposed RGBT-Tiny dataset, extensive evaluations have been conducted with IoU and SAFit metrics, including 32 recent state-of-the-art algorithms that cover four different types (\textit{i.e.,} visible generic detection, visible SOD, thermal SOD and RGBT object detection). Project is available at \url{https://github.com/XinyiYing/RGBT-Tiny}. 
\end{abstract}
\begin{IEEEkeywords}
Visible-Thermal, Tiny Object Detection, Benchmark Dataset.
\end{IEEEkeywords}}

\maketitle
\IEEEdisplaynontitleabstractindextext
\IEEEpeerreviewmaketitle
\vspace{-1cm}
\IEEEraisesectionheading{\section{Introduction}\label{sec:introduction}}
\vspace{-0.2cm}
Small objects, featured by their extremely small size (\emph{e.g.}, less than $32\times32$ pixels \cite{COCO}), are always difficult to detect. Small object detection (SOD) has received significant attention in recent years and becomes a challenging direction independent of generic object detection due to its valuable applications, including video surveillance \cite{VisDrone,Zhao2018,zhao20183d}, autonomous driving \cite{TJU-DHDTraffic,zhao2017dual,Zhao2019} and water rescue \cite{SODA,zhao2020towards,Zhao2018b,Wang2022}. Currently, the advancement of SOD faces the following challenges. First, the extremely small size with significantly fewer appearance cues raises serious limitations for feature representation learning, whereas the complex background clutter negatively affects the detection of small objects, and can cause many false alarms. Second, the lack of large-scale, high-quality datasets greatly hinders the advancement of SOD. Finally, the IOU-based evaluation metrics commonly used for generic object detection have a low tolerance for bounding box (bbox) perturbation of small objects and cannot guarantee high localization accuracy. Therefore, in this paper, we aim to address the aforementioned challenges to advance the development of SOD by firstly building a large-scale dataset for SOD, then developing a novel evaluation metric for SOD, and finally extensively evaluating various deep feature learning methods for SOD with the developed dataset and the proposed metric.

{We build a new large-scale dataset for SOD to address the following core issues.
First, most existing studies focus on either visible \cite{RFLA,QueryDet,C3Det,SAHI,WiderPerson,DOTA,SOD,TinyPerson,WiderFace,EuroCityPersons,VisDrone,SODA,TJU-DHDTraffic,SODA10M} or thermal \cite{ISNet,DNAnet,MDvsFA,CQU-SIRST,ACM,ISNet,NUDT-SIRST-Sea,SIATD,Hui1,TBC-Net,Anti-UAV,Hui2} modality independently, and few research is conducted to explore the multimodal information fusion within visible-thermal (RGBT) bimodality \cite{KAIST,MBNet,UA-CMDet}. 
Second, although various RGBT datasets \cite{Anti-UAV,Anti-UAV410,LasHeR,RGBT210,RGBT234,VTUAV,KAIST,CVC-14,LLVIP,VEDAI,Takumi,MFNet,FLIR,DVTOD,M3FD,UA-CMDet,OTCBVS,LITIV,GTOT,VOT-RGBT,QFDet} have been proposed, the insufficient quantity \cite{VOT-RGBT,DVTOD,M3FD}, limited diversity \cite{Anti-UAV410,DVTOD,M3FD}, unitary application \cite{DVTOD,M3FD,UA-CMDet}, misaligned images \cite{UA-CMDet,VEDAI,FLIR} and large target size \cite{LasHeR,VOT-RGBT,RGBT234} cannot provide an impartial benchmark for performance evaluation on RGBT SOD. 
The aforementioned issues urge us to build the first large-scale benchmark with high diversity for RGBT SOD (namely RGBT-Tiny) to advance the algorithm development and piratical applications of visible SOD, thermal SOD, RGBT image fusion, RGBT small object detection and tracking.}

{We propose a new scale adaptive fitness (SAFit) measure to guarantee robust evaluation on both large and small objects. Specifically, SAFit performs size-aware sigmoid weighted summation between large object-friendly IoU \cite{COCO} measure and small object-friendly NWD \cite{NWD} measure, which can rapidly switch to an appropriate measure according to the corresponding bbox size. The switch point is flexibly controlled by a size-aware parameter $C$ for custom requirements. In addition, a corresponding SAFit loss is developed and is demonstrated to benefit the detection performance.} 

Based on the proposed RGBT-Tiny dataset, we conducted extensive performance evaluations with IoU and SAFit metrics on 30 recent state-of-the-art algorithms, including visible generic detection \cite{SSD,YOLO,TOOD,FasterRCNN,SABL,CascadeRCNN,DynamicRCNN,RetinaNet,CenterNet,FCOS,ATSS,VarifocalNet,Deformable-DETR,SparseRCNN,CO-DETR,DiffusionDet,DINO,DDQ}, visible SOD \cite{RFLA,QueryDet,C3Det}, thermal SOD \cite{DNAnet,ALCNet,ACM}, and RGBT object detection \cite{UA-CMDet,ProbEn,QFDet,CALNet,DVTOD}. Section \ref{base} summarizes the challenges and effective schemes of RGBT-Tiny benchmark to lay a solid foundation for the research of RGBT SOD.

The main contributions of this paper can be summarized as follows: 1) We build the first large-scale benchmark dataset (namely RGBT-Tiny) with high diversity for RGBT SOD, including 115 paired sequences, 93K frames and 1.2M manual annotations. {As compared with 32 existing benchmark datasets (including visible SOD, thermal SOD, RGBT object detection and tracking datasets), RGBT-Tiny is finely aligned, and contains abundant small objects, high diversity scenes and high-quality annotations, as shown in Fig.~\ref{show}.} 2) We propose a scale adaptive fitness (SAFit) measure that exhibits high robustness to both large and small objects. The proposed SAFit can provide reasonable performance evaluation, and promote detection performance when equipped during training. 3) Based on the proposed RGBT-Tiny dataset and SAFit measure, we make comprehensive evaluations on 30 current state-of-the-art algorithms, including visible generic detection, visible SOD, thermal SOD and RGBT object detection methods, which lays solid foundations for further research.

\section{Related Work}
\label{relatedwork}

\noindent
\textbf{Visible SOD.}
Based on the current powerful paradigms \cite{FCOS,FasterRCNN,CenterNet,Deformable-DETR} on generic object detection, existing visible SOD methods \cite{SODA,Fpillars} introduce delicate designs tailored for small objects, including pre-processing \cite{DataAug1,DataAug2,ReDet,RFLA}, multi-scale representation \& fusion \cite{FPN,MSRF2,MSRF3,QueryDet}, contextual information\cite{Contextual1,Contextual2,C3Det,SAHI} and super-resolution \cite{SR1,SR2,SR3,SR4}. Among them, detection under low vision conditions (e.g., night, fog and rain) is rarely explored \cite{VisDrone,TJU-DHDTraffic}, but is significant for practical applications \cite{ISNet,Anti-UAV410}.

\noindent
{\textbf{Thermal SOD} performs robust on inferior illumination and weather conditions.
In the past decades, numerous single-frame thermal SOD methods have been proposed, including early traditional paradigms (\textit{e.g.,} filtering-based \cite{tophat,Max-Median}, local contrast-based \cite{LCM,RLCM,MSLCM,WSLCM,TLLCM}, low rank-based methods \cite{RIPT,NRAM,low_rank1,PSTNN,MSLSTIPT}) and recent deep learning paradigms \cite{MDvsFA,ACM,ALCNet,DNAnet,ISNet,Mingjin1,Mingjin2,ISTDet,RISTDnet}. Based on them, temporal information can be exploited \cite{STDetection1,STDetection2,STDetection3,STDetection4} to address blur, distortion and low-contrast. 
However, thermal SOD fails in extreme conditions with undistinguished radiation and recognition scenes that require color and texture information.}

\begin{figure}[t]
	\centering
	\includegraphics[width=\linewidth]{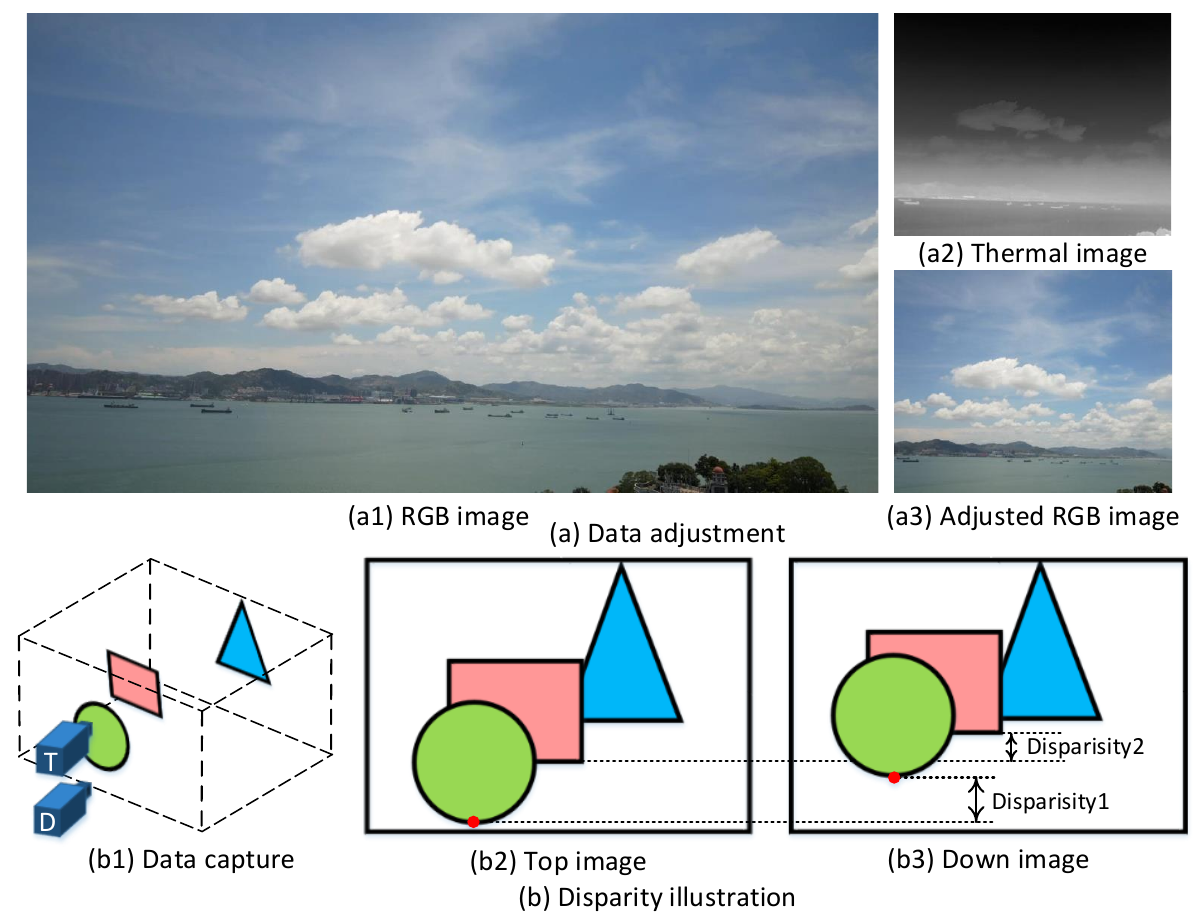}
	\caption{(a1) Raw RGB image is aligned to (a2) thermal image to generate (a3) adjusted RGB image. (b) An illustration of disparity variations of dual lenses.}\label{dual_lens} 
\end{figure}

\noindent
{\textbf{RGBT Object Detection} employ multimodal advantages for improved robustness. However, existing RGBT object detection methods always focus on specific tasks (\textit{e.g.,} pedestrian \cite{ACF,MBNet,QFDet} and vehicle \cite{UA-CMDet,Takumi,vehicle1} detection) to explore the effective multimodal fusion \cite{KAIST,MBNet,bimodality3,zhang2020multispectral,Zhang2024fusion} for spatio-temporal misalignment \cite{limultispectral,zhang2019weakly,MBNet,kim2022towards,DVTOD,Misalignment} and low light vision \cite{guan2019fusion,zhang2023illumination,MBNet}. Besides these issues, RGBT SOD exhibits more challenges, including extremely small object size, fewer appearance clues, easily interfered by clutters.}

\begin{table*}[htbp]
	\centering
	\vspace{-.03cm}
	\caption{{Statistical comparison among existing RGB SOD datasets (RGB-SOD), thermal SOD datasets (T-SOD), RGBT tracking datasets (RGBT-T), RGBT detection datasets (RGBT-D) and our RGBT-Tiny dataset. ``Seq.", ``Frame", ``Anno.", ``T-Cat." and ``S-Cat." represent the number of sequences, frames, annotations, target \& scene categories, respectively. ``FPS" is frame per second of released video sequences. ``Split" represents the way of data split. ``Align" represents whether RGBT images are aligned (Y) or not (N). ``ID" represents providing tracking ID (Y) or not (N). ``Pub" and ``Year" represent publication name and year.}}\label{tab-Statics}
	
	\setlength{\tabcolsep}{1.7mm}
	\renewcommand\arraystretch{1.1}
	\begin{tabular}{|clcccccccccccc|}
		\hline
		&Benchmark&Seq.&Frame&Anno.&Resolution&FPS&T-Cat.&S-Cat.&Split&Align&ID&Pub&Year\\\hline
		\multirow{11}*{\rotatebox{90}{RGB-SOD}}
		&{SOD4SB \cite{SOD4SB}}&{-}&{39K}&{137K}&{3840$\times$2160}&{-}&{4}&{4}&{train/test-a/test-b}&{-}&{-}&{CVPR}&{2023}\\
		&SODA-D \cite{SODA}&-&24K&278K&3407$\times$2470&-&9&-&train/val/test&-&-&TPAMI&2023\\
		&SODA-A \cite{SODA}&-&2.5K&800K&4761$\times$2777&-&9&-&train/val/test&-&-&TPAMI&2023\\
		&TJU-DHD Traffic \cite{TJU-DHDTraffic}&-&60K&332K&1624$\times$1200&-&5&-&train/val&-&-&TIP&2021\\
		&SODA-10M \cite{SODA10M}&-&10K&13K&1920$\times$1080&-&6&-&train/val&-&-&NeurIPS&2021\\
		&VisDrone \cite{VisDrone}&-&40&183K&3840$\times$2160&-&10&-&train/val/test-c/test-d&-&-&TPAMI&2021\\
		&DOTA v2 \cite{DOTA}&-&11k&1.8M&800$^2$-4000$^2$&-&18&-&train/val/test-c/test-d&-&-&TPAMI&2021\\
		&WiderPerson \cite{WiderPerson}&-&13k&400K&Varied&-&1&-&train/val/test&-&-&TMM&2020\\
		&EuroCity Persons \cite{EuroCityPersons}&-&47k&211K&1920$\times$1024&-&3&-&train/val/test&-&-&TPAMI&2020\\
		&WiderFace \cite{WiderFace}&-&32K&394K&Varied&-&1&61&train/val/test&-&-&CVPR&2016\\
		&MS-COCO \cite{COCO}&-&328k&2.5M&Varied&-&91&-&train/val/test&-&-&ECCV&2014\\ \hline
		\multirow{7}*{\rotatebox{90}{T-SOD}}
		&{SIRST-v2 \cite{OSCAR}} &{-}&{514}&{648}&{278$\times$366}&{-}&{1}&{3}&{train/val/test}&{-}&{-}&{TGRS}&{2023}\\
		&IRSTD-1K \cite{ISNet}&-&1K&1.5K&512$\times$512&-&1&6&train/val/test&-&-&CVPR&2022\\
		&NUDT-SIRST \cite{DNAnet}&-&1.3K&1.9K&256$\times$256&-&1&5&train/test&-&-&TIP&2022\\
		&{IRDST \cite{RDIAN}}&{401}&{147K}&{144K}&{795$\times$553}&{-}&{1}&{-}&{train/test}&{-}&{-}&TGRS&2023\\
		&Fu \cite{Hui2}&87&22K&89K&640$\times$512&35&1&2&train/val&-&-&CSD&2022\\
		&SIATD \cite{SIATD}&350&150K&247K&640$\times$512&30&1&3&train/test&-&-&CSD&2021\\
		&Hui \cite{Hui1}&22&16K&17K&256$\times$256&100&2&-&-&-&-&CSD&2020\\\hline
		\multirow{5}*{\rotatebox{90}{RGBT-T}}
		&{Anti-UAV410 \cite{Anti-UAV410}}&{410}&{438K}&{438K}&{640$\times$512}&{25}&{1}&{7}&{train/val/test}&{N}&{N}&{TPAMI}&{2024}\\
		&VTUAV \cite{VTUAV}&500&3.3M&326K&1920$\times$1080&30&13&15&train/test&Y&N&CVPR&2022\\
		&LasHeR \cite{LasHeR}&1224&6.7M&6.7M&630$\times$480&-&32&20+&train/test&Y&N&TIP&2021\\
		&VOT-RGBT \cite{VOT-RGBT}&60&40K&40K&630$\times$460&20&13&-&-&Y&N&ICCVW&2020\\
		&RGBT234 \cite{RGBT234}&234&234K&234K&630$\times$460&30&22&-&-&Y&N&PR&2019\\ \hline
		\multirow{9}*{\rotatebox{90}{RGBT-D}}&{DVTOD \cite{DVTOD}}&{-}&{4.4K}&{12K}&{1920$\times$1080}&{-}&{3}&{-}&{-}&{N}&{-}&{TIV}&{2024}\\ 
		&{RGBTDronePerson \cite{QFDet}}&{-}&{16.8K}&{74K}&{640$\times$512}&{-}&{1}&{-}&{train/test}&{N}&{-}&{ISPRS}&{2023}\\ 
		&{M$^3$FD \cite{M3FD}}&{-}&{8.4K}&{34K}&{1024$\times$768}&{-}&{6}&{8}&{-}&{Y}&{-}&{CVPR}&{2022}\\ 
		&{DroneVehicle \cite{UA-CMDet}}&{-}&{57K}&{953K}&{840$\times$712} &{-}&{5}&{-}&{train/val/test}&{N}&{-}&{TCSVT}&{2021}\\
		&LLVIP \cite{LLVIP}&-&34K&-&1080$\times$720&-&1&-&-&Y&-&ICCV&2021\\ 
		&VEDAI \cite{VEDAI}&-&1.2K&3.7K&512$^2$-1024$^2$&-&9&-&train/test&N&-&JVCI&2016\\ 
		&CVC-14 \cite{CVC-14}&4&17K&18K&640$\times$512&10&1&-&train/test&Y&-&Sensors&2016\\ 
		&KAIST \cite{KAIST}&41&191K&103K&640$\times$480&20&3&-&train/test&Y&-&CVPR&2015\\ 
		&FLIR \cite{FLIR}&7498&26K&520K&640$\times$512&24&15&-&train/val&N&-&-&2018\\ \hline
		\cellcolor{mygray}{}&\cellcolor{mygray}{RGBT-Tiny} &\cellcolor{mygray}{115}&\cellcolor{mygray}{93K}&\cellcolor{mygray}{1.2M}&\cellcolor{mygray}{640$\times$512}&\cellcolor{mygray}{15}&\cellcolor{mygray}{7}&\cellcolor{mygray}{8}&\cellcolor{mygray}{train/test}&\cellcolor{mygray}{Y}&\cellcolor{mygray}{Y}&\cellcolor{mygray}{-}&\cellcolor{mygray}{2024}\\\hline
	\end{tabular}
\end{table*}

\noindent
\textbf{RGBT Datasets.} Early RGBT datasets \cite{OTCBVS,LITIV,GTOT,VOT-RGBT} cannot satisfy the ``data hunger" of recent deep learning-based methods due to limited quantity and category. Then large-scale datasets \cite{Anti-UAV,Anti-UAV410,LasHeR,RGBT210,RGBT234,VTUAV,KAIST,CVC-14,LLVIP,VEDAI,Takumi,MFNet,FLIR,DVTOD,M3FD,UA-CMDet} with abundant objects \& scenes and various applications have been proposed. However, the insufficient quantity \cite{VOT-RGBT,DVTOD,M3FD,UA-CMDet,LLVIP,VEDAI,CVC-14,FLIR}, limited diversity \cite{Anti-UAV410,DVTOD,M3FD,UA-CMDet,LLVIP,VEDAI,CVC-14}, unitary application \cite{DVTOD,M3FD,UA-CMDet,LLVIP,VEDAI,CVC-14,FLIR,Anti-UAV410,VTUAV,LasHeR,VOT-RGBT,RGBT234}, misaligned images \cite{Anti-UAV410,DVTOD,UA-CMDet,VEDAI,FLIR} and large target size \cite{DVTOD,M3FD,UA-CMDet,LLVIP,VEDAI,CVC-14,FLIR,Anti-UAV410,VTUAV,LasHeR,VOT-RGBT,RGBT234} cannot provide an impartial benchmark to evaluate RGBT SOD algorithms. The aforementioned issues urge us to build the first large-scale benchmark dataset RGBT-Tiny. Recently, Xu \textit{et. al.} \cite{MOT-QY} further developed the corresponding MOT labels to promote the research development of RGBT tiny object tracking.

\noindent
\textbf{Evaluation Metrics.} Intersection over union (IoU), average precision (AP) and average recall (AR) based on bbox are widely used evaluation metrics of visible SOD \cite{SODA,Fpillars}. Based on IoU, numerous modified versions have been proposed, including generalized IoU (GIoU) \cite{GIoU}, distance IoU (DIoU) \cite{DIoU} and complete-IoU (CIoU) \cite{DIoU}. However, these metrics focus on non-overlapping bboxes, but cannot well address the inherent problem of low tolerance for bbox perturbation. Therefore, a novel evaluation metric tailored for SOD is absolutely necessary. 

\section{RGBT-Tiny Benchmark}

\subsection{Data Collection and Annotations}

\noindent
\textbf{Data Capture.}
We employ a professional UAV DJI Mavic 2 as the data acquisition platform to ensure stable flight in extreme conditions. Vertically arranged RGBT dual lenses are equipped in UAV to collect RGBT video sequences from an altitude of 60-100 meters. The frame rate of visible and thermal cameras is 30, and we sample the video sequence to 15 frames per second (FPS) in public videos for more obvious temporal motion. Thermal camera has a wavelength of 8-14 $\mu m$, and the image sizes between visible and thermal cameras are different (\textit{i.e.,} 1080$\times$1920 of RGB images and 512$\times$640 of thermal images). 

\noindent
\textbf{Data Adjustment.}
Camera calibration \cite{calibration} is first applied to remove lens distortion in RGBT images. Then, we employ homography transformation \cite{homography} to align RGB images to thermal images since the positions of RGBT cameras are relatively fixed. 
To address the resolution difference between RGBT images, we crop the aligned RGB image patches that are consistent with thermal images to generate paired RGBT images with a resolution of 640$\times$512. The adjusted RGBT images are shown in Fig.~\ref{dual_lens} (a). Note that, homography transformation can only perform frame alignment within a fixed depth of field (DoF). Therefore, the inherent disparity variation (shown in Fig.~\ref{dual_lens} (b)) of dual lenses \cite{disparities} has not been well solved, and is a challenge that deserves investigation.

\begin{figure}[t]
	\centering
	\includegraphics[width=\linewidth]{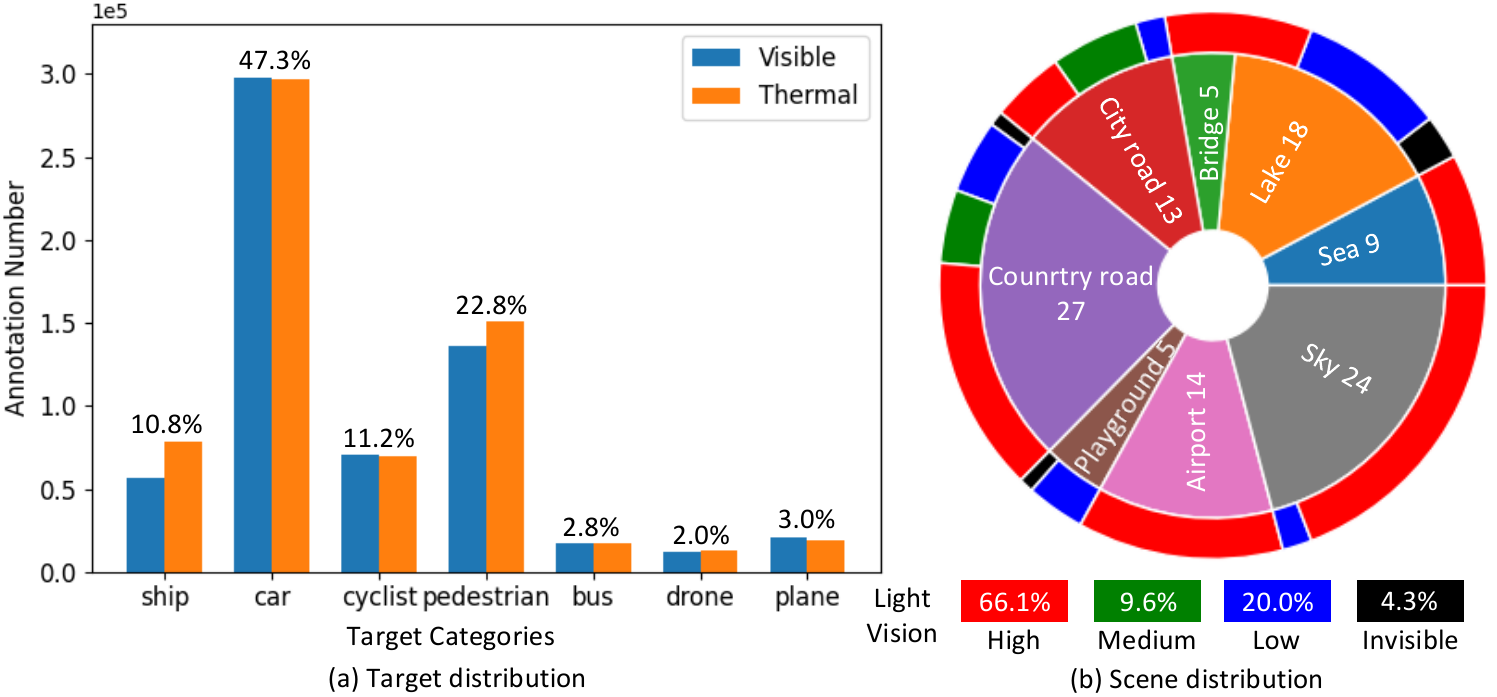}
	\caption{(a) Annotation numbers w.r.t. target categories in visible and thermal modalities. Numbers represent the proportion of each category in annotations. (b) Inner circle shows sequence numbers w.r.t. scene categories, and outer circle shows the light vision distribution of scenes. Numbers in the pie chart represent the number of sequences of each scene type. Numbers in the legend represent the proportion of each light vision in annotations.}\label{target_scenes} 
\end{figure}

\noindent
\textbf{Groundtruth Annotations.} 
We use DarkLabel \cite{DarkLabel} to annotate the groundtruth (GT) bbox with corresponding category and tracking ID. Note that, except for a few unrecognizable annotations in extreme conditions, RGBT annotations are paired within one-to-one correspondence. 
For quality assurance, we spent over 2000 hours conducting two-step verification. 1) 
Ten professional annotators make annotations respectively, and review one other annotator. 2) Each image is evaluated by another 2 assessors (5 assessors in total), and annotations are constantly revisited until without skepticism.

\noindent
\textbf{Training and Test Sets.} To avoid data bias and over-fitting, training and test sets are split for 85 and 30 video sequences by the following criterion. 1) Each subset covers all types of scenes and objects. 2) Each subset covers all illumination and density variations. 3) Two subsets are not overlapped. 

\subsection{Benchmark Properties and Statistics}

\noindent
\textbf{Rich Diversity.} As shown in Fig.~\ref{target_scenes} (a), objects can be divided into 7 categories (\textit{e.g.,} ship, car, cyclist, pedestrian, bus, drone and plane). It can be observed that despite generally consistent, the number of annotations in thermal images is higher (\textit{e.g.,} ship and pedestrian) than those in visible images. This is because, as shown in Fig.~\ref{target_scenes} (b), our dataset covers different light visions (\textit{i.e.,} high-light vision is captured in the daytime, and medium-light, low-light \& invisible visions are captured at night), and thermal images can provide additional supplemental information under low-light and invisible visions. Note that, night-time sequences occupy 33.9\% of all data, and over 70\% of them are in low-light and invisible visions. 
Sequences are captured at 8 types of scenes (\textit{e.g.,} sea, lake, bridge, city road, country road, playground, airport and sky) across four cities over a period of one year to obtain data in different seasons, weather and locations. {Comparisons among existing datasets are listed in Table~\ref{tab-Statics}. In conclusion, we provide the first, large-scale, finely-aligned RGBT SOD datasets with abundant objects \& scenes and high-quality annotations, which facilitates the development of RGBT image fusion, object detection and tracking.}

\begin{figure}[t]
	\centering
	\includegraphics[width=\linewidth]{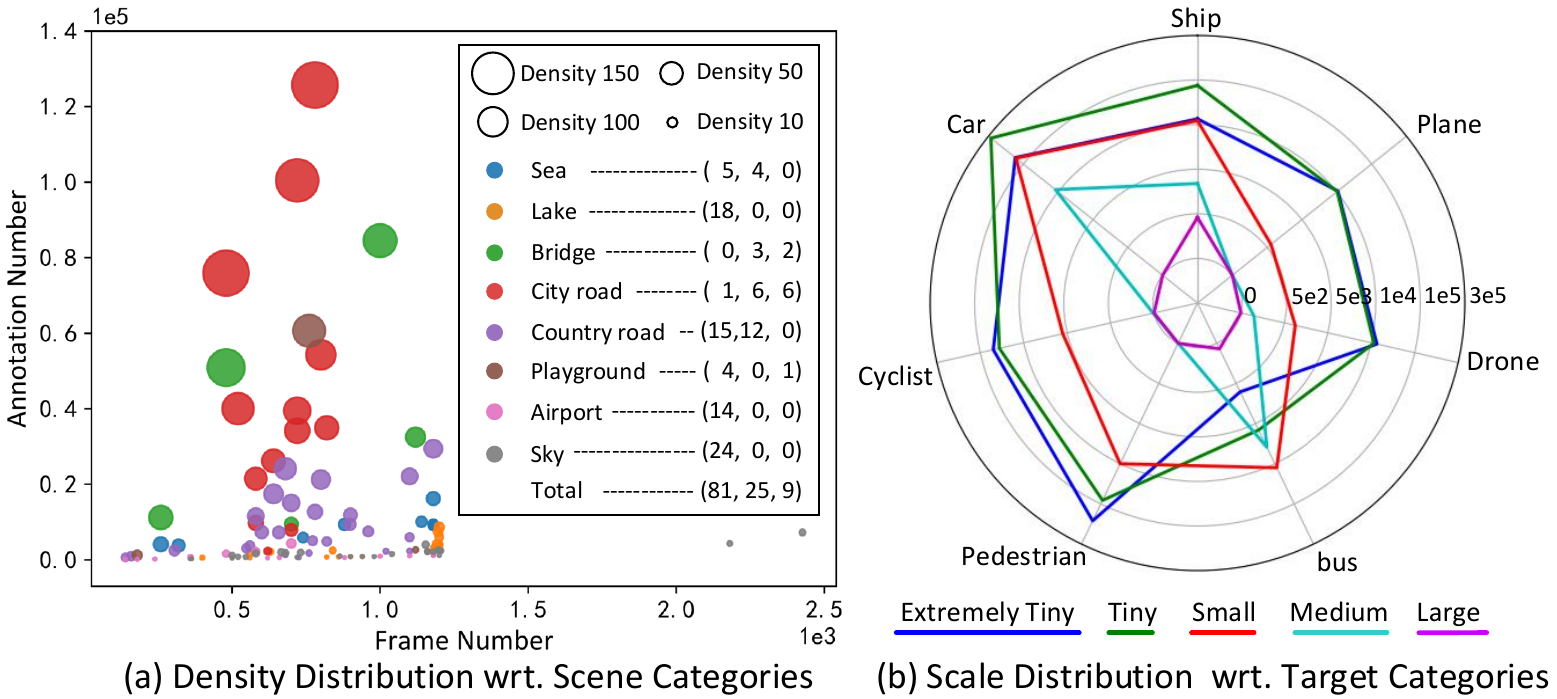}
	\caption{(a) Average annotation number per frame (\textit{i.e.,} annotation density) of each sequence. Larger circle represents higher density, and different colors represent different scene types. ($x$,$y$,$z$) are the numbers of sequences w.r.t. density levels (\textit{i.e.,} sparse, medium, dense). (b) Size distribution of each target category. Lines with different colors represent different scale levels. Radius represents the annotation number, and the area under each color line represents the total annotation number of each scale level.}\label{density_scale}
	\vspace{-.45cm}
\end{figure}

\begin{figure*}[t]
	\centering
	\includegraphics[width=1\textwidth]{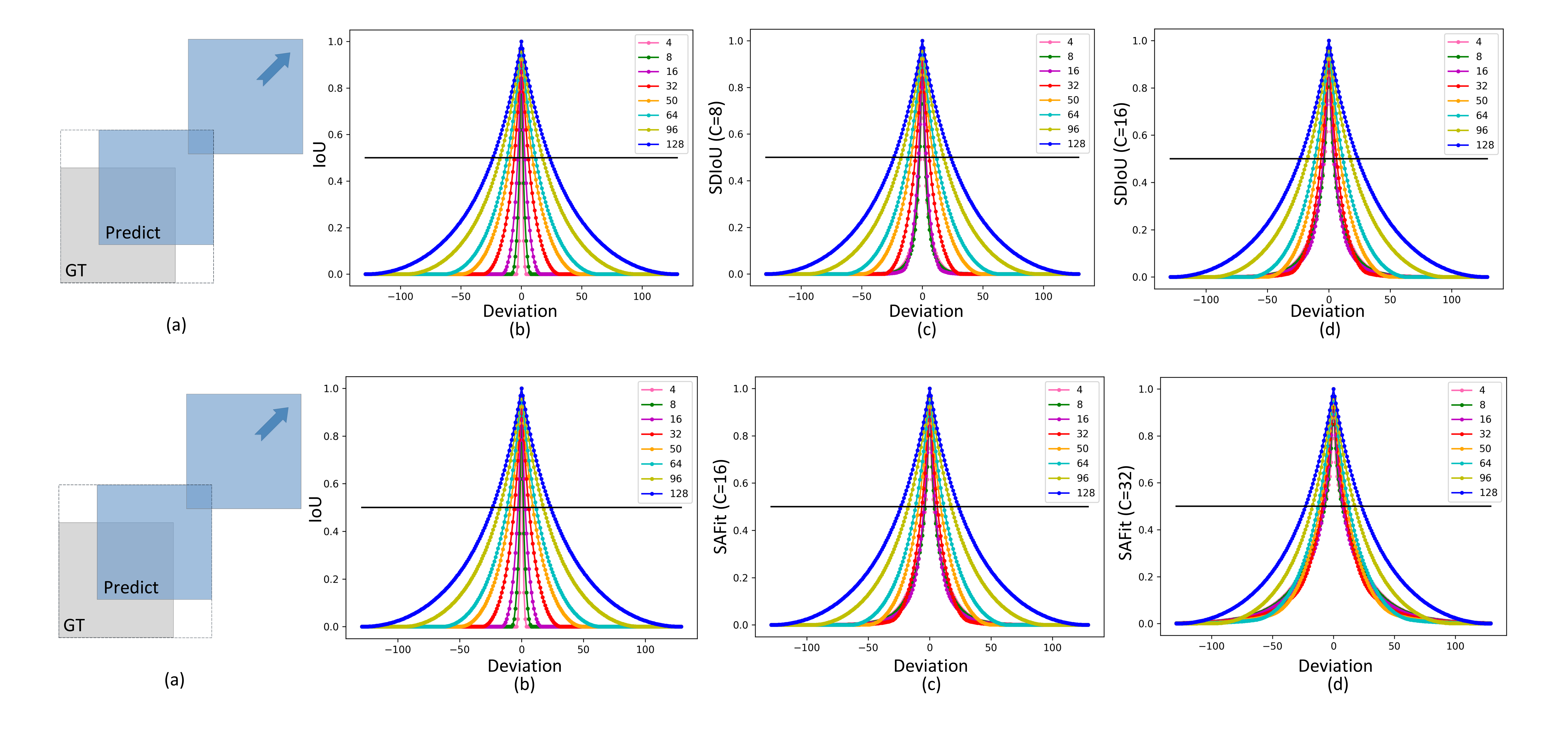}
	\caption{{(a) An illustration of the pixel deviation between the center points of GT bbox and predicted bbox. (b) IoU-Deviation curves w.r.t different sizes of bboxes. (c)-(d) SAFit-Deviation curves under different $C$ values. The abscissa value represents the number of pixels deviation. The ordinate value represents the corresponding metric value. Note that, since the locations of bboxes can only change discretely, curves are presented as scatter diagrams.}}\label{fig_evaluation}
	\vspace{-.4cm}
\end{figure*}

\noindent
\textbf{Large Density Variations.} Fig.~\ref{density_scale} (a) shows the average annotation number per frame (\textit{i.e.,} annotation density) of each sequence, and we divide density into three levels: sparse$\in$[1,10), medium$\in$[10,50), dense$\in$[50,$\infty$). It can be observed that our dataset covers a large range of annotation density (from 1 to 161), and density varies greatly among different scenes. Specifically, the density of city roads and bridges is much higher than that of sky and airports due to their unique objects and applications, which can provide valuable priors for object detection and recognition. 

\noindent
\textbf{Small-Scale Objects.} Following general scale ranks\footnote{small$\in$[$1^2$,$32^2$), medium$\in$[$32^2$,$96^2$) and large$\in$[$96^2$,$\infty$)} \cite{COCO}, we further divide the small scale into three levels: extremely tiny$\in$[$1^2$,$8^2$), tiny$\in$[$8^2$,$16^2$), small$\in$[$16^2$,$32^2$). Fig.~\ref{density_scale} (b) shows the annotation number with respect to (w.r.t.) the scale of each target category. It can be observed that tiny objects occupy the largest proportion (\textit{i.e.,} 48\%) and over 97\% of objects are within small or smaller scales. In addition, due to different angles (\textit{i.e.,} up, front and down) and distance of data acquisition, the absolute target size in image is different from the real target size. For example, larger planes are mostly divided into extremely tiny scale due to far-front handheld capture conditions, and smaller buses are mostly divided into small scale due to close-down flight capture conditions. In conclusion, objects cannot be simply classified by their absolute sizes. Comprehensive properties including appearance, context, density and motion information should be considered for accurate detection.

\noindent
\textbf{Temporal Occlusion.} For short-time occlusion (less than 5 frames), we employ temporal interpolation of bboxes \cite{DarkLabel,Xie2024} to maintain consistency. For long-time occlusion (more than 5 frames), occluded frames remain unsolved. Among all annotations, 3.4\% are slightly occluded (5-10 frames), 3.4\% are moderately occluded (10-20 frames) and 5.2\% are heavily occluded (more than 20 frames).

\subsection{Scale Adaptive Fitness Measure}

{Normalized Wasserstein distance (NWD) \cite{NWD} has been demonstrated to be friendly to SOD due to scale invariance and smoothness to location deviation. The formulation can be defined as:
\vspace{-.2cm}
\begin{align}\scriptsize
	\rm{NWD}(K)&=\exp \left(-\frac{\sqrt{W_2^2\left(\mathcal{N}_{p}, \mathcal{N}_{gt}\right)}}{K}\right),
\end{align}	
\vspace{-.4cm}
\begin{align}\scriptsize
	W_2^2\left(\mathcal{N}_p, \mathcal{N}_{gt}\right)=\left\|\left(\mathcal{N}_p^{\mathrm{T}},\mathcal{N}_{gt}^{\mathrm{T}}\right)\right\|_2^2,
\end{align}
where $W_2^2\left(\mathcal{N}_p, \mathcal{N}_{gt}\right)$ is the Wasserstein distance between the Gaussian distributions of predicted bbox $\mathcal{N}_p$ = $\left[c x_p, c y_p, w_p/2, h_p/2\right]$ and GT bbox $\mathcal{N}_{gt}$ = $\left[c x_{gt}, c y_{gt}, w_{gt}/2, h_{gt}/2\right]$ with center point locations of $(cx, cy)$, width $w$ and height $h$. $K$ is a hyperparameter closely related to the dataset \cite{NWD}. However, the scale-invariant absolute distance measure cannot provide reasonable evaluation for objects with large sizes.}

{Intersection over union (IoU) \cite{Fpillars} is a common and reasonable metric for performance evaluation on large generic objects. The formulation can be defined as:
\vspace{-.1cm}
\begin{align}
	\rm{IoU}&= \frac{S_{p}\cap S_{gt}}{S_{p}\cup S_{gt}},
\end{align}
where $S_{p}$ and $S_{gt}$ represent the predicted and GT bboxes. Despite scale invariance, symmetry and other advantages, IoU shows low tolerance for bbox perturbation of small objects, as shown in Fig.~\ref{fig_evaluation} (b). A minor location deviation (\textit{e.g.,} 2 pixels deviation for a small object of size 8$\times$8) can lead to a notable IoU drop (\textit{e.g.,} from 1 to 0.39). In conclusion, IoU-based metrics are not suitable for evaluating the performance of SOD.}

{To combine both advantages of IoU and NWD while avoiding drawbacks, we develop a scale adaptive fitness (SAFit) measure that exhibits high robustness to both large and small objects. Specifically, we combine IoU and NWD via size-aware Sigmoid weighted summation:
\begin{equation}
	\begin{aligned}
		\rm{SAFit}&=\frac{1}{1+e^{-(\sqrt{A}/C-1)}} \times \rm{IoU} \\
		&+(1-\frac{1}{1+e^{-(\sqrt{A}/C-1)}}) \times \rm{NWD}(C),
	\end{aligned}
\end{equation}
where Sigmoid function indicates a soft switch, which can rapidly switch to an appropriate measure by corresponding bbox size. $A$ is the area of GT bbox and $C$ is a constant that balances NWD and IoU measures in a size-aware manner. That is, when $A$ = $C^2$, NWD and IoU share equal contributions. Lower value of $A$ (\textit{i.e.,} smaller size of GT bbox) leads to NWD domination, while higher value results in an increased proportion of IoU. In conclusion, SAFit is practical for real applications that contain objects with varied sizes.}

Quantitative comparisons among IoU and SAFit under different $C$ values (\textit{i.e.,} 16, 32) are shown in Fig.~\ref{fig_evaluation}. It can be observed that when the size of GT bbox is larger than $C$, SAFit is consistent with IoU. As GT bbox size decreases, SAFit rapidly turns to NWD, which is highly robust to bbox perturbation. Note that, through adjusting the value of $C$, SAFit can provide flexible applications for different custom requirements. For our dataset, we set $C$ = 32 because small target is defined to be smaller than 32$\times$32. In addition, we develop SAFit loss (\textit{i.e.,} $L_{\rm{SAFit}}= 1-\rm{SAFit}$) for network training, which can provide stable and accurate optimization guidance on objects with varied sizes. Note that, each component (\textit{i.e.,} IoU and NWD) of SAFit loss can be flexibly replaced by new measures.

\section{Experiments}

\subsection{Scale Adaptive Fitness Measure}

\noindent
\textbf{SAFit Measure for Evaluation.}
{We employ IoU, NWD and SAFit measures for performance evaluation on Cascade RCNN \cite{CascadeRCNN} in visible modality. More results are listed in supplemental material. As shown in Fig.~\ref{fig_dis_iou} (a), AP values of SAFit are closer to those of NWD when GT bbox size is small, and rapidly switch to those of IoU as size increases.} In conclusion, SAFit shows comprehensively reasonable evaluations on both large and small objects. To this end, all experimental results below are evaluated under SAFit measure if not specified.

\noindent
\textbf{SAFit Loss for Training.} We equip different losses (\textit{i.e.,} IoU \cite{Fpillars}, DIoU \cite{DIoU}, CIoU \cite{DIoU}, GIoU \cite{GIoU}, NWD \cite{NWD} and SAFit losses) with different detectors (\textit{i.e.,} ATSS \cite{ATSS}, SparseRCNN \cite{SparseRCNN}, and train the network under the same settings in visible modality. Note that, we employ 2 variants of SAFit loss (\textit{i.e.,} SAFit-s and SAFit$_g$) to investigate the performance of direct transition (\textit{i.e.,} loss function is set to NWD when GT bbox size is smaller than $C$, while set to IoU vice verse) and stronger component (\textit{i.e.,} Sigmoid weighted summation of GIoU and NWD). SAFit-based results are shown in Table~\ref{tab-losses}. Please refer to supplemental material for corresponding IoU-based results. It can be observed that SAFit loss shows high robustness to different detectors under SAFit and IoU metrics. Compared with SAFit, SAFit-s performs inferior while SAFit$_g$ achieves improved performance, which demonstrates the superiority and flexibility of SAFit. Furthermore, SAFit$_g$ and SAFit achieve higher values compared with their components, which demonstrates that SAFit can not only combine both advantages of their components but also produce a breakthrough by offering more stable and smooth training for objects with varied sizes.

\begin{figure}[t]
	\centering
	\vspace{-.05cm}
	\includegraphics[width=\linewidth]{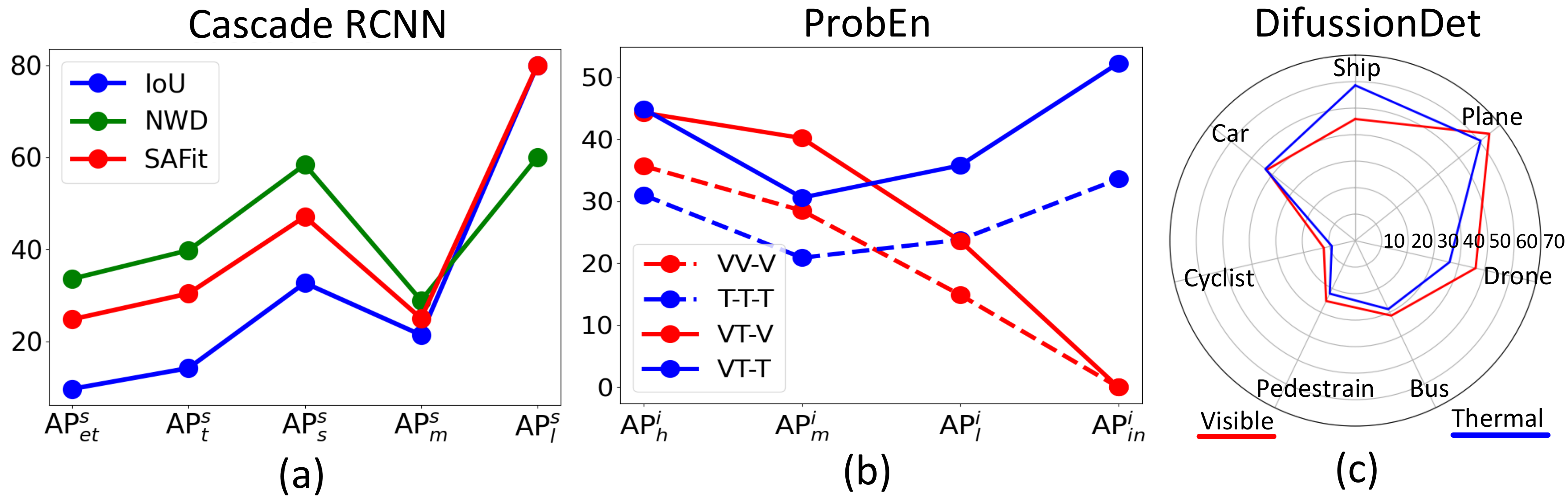}
	\vspace{-.65cm}
	\caption{{(a) shows comparisons among different measures for performance evaluation. AP$_{et}^s$, AP$_{t}^s$, AP$_s^s$, AP$_m^s$, AP$_l^s$ represent AP values of extremely tiny, tiny, small, medium, large objects. (b) investigates the influences of RGBT fusion. ``VV", ``TT", ``VT" represent model with duplicated visible, duplicated thermal and RGBT inputs. ``-V", ``-T" represent visible, thermal labels for training and test. AP$_h^i$, AP$_m^i$, AP$_l^i$, AP$_{in}^i$ represent AP values under high-light, medium-light, low-light, invisible illumination conditions. (e) shows performance comparisons of different classes.}}\label{fig_dis_iou} 
	\vspace{-.33cm}
\end{figure}

\subsection{Baseline Results}\label{base}
{We conduct comprehensive evaluations on 30 recent state-of-the-art detection methods, including 18 visible generic object detection methods\footnote{{All models are implemented by mmdetection code library \cite{mmdetection} under their default parameters (ResNet50 \cite{ResNet50} and FPN \cite{FPN} are preferred as the backbone and neck) and training settings. We reduce the initial anchor size of two-stage methods to adapt to small objects.}} (\textit{e.g.,} SSD \cite{SSD}, YOLO \cite{YOLO}, TOOD \cite{TOOD}, Faster RCNN \cite{FasterRCNN}, SABL \cite{SABL}, Cascade RCNN \cite{CascadeRCNN}, Dynamic RCNN \cite{DynamicRCNN}, RetinaNet \cite{ RetinaNet}, CenterNet \cite{CenterNet}, FCOS \cite{FCOS}, ATSS \cite{ATSS}, VarifocalNet \cite{VarifocalNet}, Deformable DETR \cite{Deformable-DETR} Sparse RCNN \cite{SparseRCNN}, CO-DETR \cite{CO-DETR}, DiffusionDet \cite{DiffusionDet}, DINO \cite{DINO} and DDQ \cite{DDQ}), 3 visible SOD methods\footnote{{All models are implemented by their officially public codes.}\label{web}} (\textit{e.g.,} RFLA \cite{RFLA}, QueryDet \cite{QueryDet}, C3Det \cite{C3Det}), 3 thermal SOD methods\footnote{{All models are implemented by BasicIRSTD code library \cite{BasicIRSTD} under their default settings.}\label{web1}} (\textit{e.g.,} ACM \cite{ACM}, ALCNet \cite{ALCNet}, DNA-Net \cite{DNAnet}) and 6 RGBT detection methods\textsuperscript{\ref {web}} (\textit{e.g.,} UA-CMDet \cite{UA-CMDet}, ProbEn-early \cite{ProbEn}, ProbEn-middle \cite{ProbEn}, QFDet \cite{QFDet}, CALNet \cite{CALNet}, CMA-Det \cite{DVTOD}). SAFit-based results are shown in tables~\ref{tab-Algorithms}, and IoU-based results are listed in the supplemental material. Note that, all models are re-trained and evaluated on RGBT-Tiny dataset for fair comparisons.} 

\begin{table}
	\scriptsize
	\centering
	\vspace{0.1cm}
	\caption{SAFit-based results of different losses equipped with different detectors. SAFit-s, SAFit$_g$ are used to investigate the effect of direct transition (\textit{i.e.,} loss function is set to NWD when GT bbox size is smaller than $C$, while set to IoU vice verse) and stronger components (\textit{i.e.,} Sigmoid weighted summation of GIoU and NWD). 
	}\label{tab-losses}
	\setlength{\tabcolsep}{0.28mm}
	\renewcommand\arraystretch{.95}
	\vspace{-0.2cm}
	\begin{tabular}{|l|cc|ccccc|cc|ccccc|}
		\hline
		\multicolumn{1}{|c}{\multirow{2}*{Loss}}&\multicolumn{7}{|c|}{ATSS}&\multicolumn{7}{|c|}{Sparse RCNN}\\\cline{2-15}
		&AP&AP$_{50}$&AP$_{et}^s$&AP$_{t}^s$&AP$_s^s$&AP$_m^s$&AP$_l^s$&AP&AP$_{50}$&AP$_{et}^s$&AP$_{t}^s$&AP$_s^s$&AP$_m^s$&AP$_l^s$\\\hline
		GIoU \cite{GIoU}&24.2&38.1 &19.6 &23.6 &43.8 &27.5 &65.0 &19.2&29.8 &18.8 &\underline{19.5} &33.4 &14.9 &40.1 \\
		DIoU \cite{DIoU}&23.6&37.5 &19.2 &23.1 &43.6 &\underline{28.2} &\textbf{80.0} &19.1&29.7 &19.7 &18.5 &33.9 &14.7 &\underline{60.1} \\
		CIoU \cite{DIoU}&24.2&39.1 &19.6 &24.2 &42.6 &\textbf{29.0} &65.0 &20.0&30.6 &19.5 &19.0 &33.9 &14.8 &\textbf{80.1} \\
		IoU \cite{COCO}&23.3&37.3 &19.3 &23.7 &42.3 &26.5 &\underline{75.0} &8.1&14.7 &11.3 &7.1 &11.8 &0.2 &0.0 \\
		NWD \cite{NWD}&24.1&38.6&19.6&23.6&\underline{44.2}&27.0&60.0&19.7&30.2 &19.7 &18.4 &\underline{35.8} &\underline{15.2} &50.0 \\
		SAFit-s&24.3&39.0&\underline{19.7}&\underline{24.3}&42.1&27.5&60.0&19.8&31.2&18.4&19.4&35.5&14.5&50.1 \\
		SAFit&\underline{24.5}&\underline{39.2}&\underline{19.7}&\textbf{25.1}&43.5&28.0&60.1 &\underline{21.4}&\underline{32.2}&\underline{20.4} &\textbf{20.5} &\textbf{36.3} &14.5 &15.0 \\
		SAFit$_g$&\textbf{24.7}&\textbf{39.3}&\textbf{20.6}&23.6&\textbf{44.4}&27.5&70.0&\textbf{22.0}&\textbf{34.0} &\textbf{22.2} &\underline{19.5}&34.9&\textbf{16.3} &15.0 \\
		\hline
	\end{tabular}
	\vspace{-0.25cm}
\end{table}

\begin{table*}
	\footnotesize
	\centering
	\caption{{SAFit-based results of existing visible generic detection (V-D), visible SOD (V-SOD), thermal SOD (T-SOD), RGBT detection methods (RGBT-D) methods on RGBT-Tiny dataset. ``\#Param." represents the number of parameters. ``(-)" represents network trained under replicated visible or thermal inputs to investigate the influence of multimodal fusion. Note that, the results of T-SOD are trained with hard (left) and soft (right) masks generated by bboxes under uniform and Gaussian distributions.}}\label{tab-Algorithms}
	\vspace{-.2cm}
	\setlength{\tabcolsep}{1.05mm}
		\renewcommand\arraystretch{1.1}
	\begin{tabular}{|c|l|c|ccc|ccccc|c|ccc|ccccc|c|}
		\hline
		&\multicolumn{1}{c|}{\multirow{2}*{Methods}}&\multirow{2}*{\#Param.}&\multicolumn{9}{c|}{Visible}&\multicolumn{9}{c|}{Thermal}\\\cline{4-21}
		&&&AP&AP$_{50}$&AP$_{75}$&AP$_{et}^s$&AP$_{t}^s$&AP$_s^s$&AP$_m^s$&AP$_l^s$&AR&AP&AP$_{50}$&AP$_{75}$&AP$_{et}^s$&AP$_{t}^s$&AP$_s^s$&AP$_m^s$&AP$_l^s$&AR\\\hline
		\multirow{18}*{\rotatebox{90}{V-D}}&SSD \cite{SSD}&25.2M&28.0 &43.1 &31.9 &24.2 &26.6 &41.9 &22.7 &45.0 &36.8 &27.0 &42.0 &31.8 &20.2 &29.5 &45.7 &34.8 &30.0 &35.9 \\
		&YOLO \cite{YOLO}&61.5M&24.3 &37.7 &28.4 &21.4 &25.1 &35.8 &20.5 &60.1 &30.5 &24.1 &36.7 &28.4 &18.5 &29.5 &40.2 &26.3 &50.0 &30.9 \\
		&TOOD \cite{TOOD}&31.8M&27.9 &43.5 &31.7 &23.1 &27.9 &44.6 &30.6 &60.0 &38.6 &27.9 &42.3 &32.0 &22.1 &27.9 &47.9 &27.7 &70.0 &38.9 \\
		&ATSS \cite{ATSS}&31.9M&24.2 &38.1 &26.8 &19.6 &23.6 &43.8 &27.5 &65.0 &38.0 &27.5 &42.1 &32.3 &19.9 &29.7 &46.5 &44.1 &{80.0} &40.0 \\
		&RetinaNet \cite{RetinaNet}&36.2M&21.8 &37.4 &22.9 &20.9 &19.4 &34.9 &25.3 &75.0 &34.5 &19.3 &32.4 &21.7 &15.4 &21.4 &33.8 &35.7 &{80.0} &35.1 \\
		&Faster RCNN \cite{FasterRCNN}&41.2M&28.8 &43.1 &33.5 &24.3 &30.1 &44.2 &22.0 &65.0 &37.2 &29.5 &43.0 &{36.2} &21.9 &35.2 &45.0 &41.4 &{80.0} &36.4 \\
		&Cascade RCNN \cite{CascadeRCNN}&68.9M&{30.1} &44.2 &{35.8} &{24.}8 &30.4 &{47.1} &25.0 &80.0 &37.4 &30.0 &{44.2}&34.9 &{24.7} &29.8 &46.6 &25.8 &{\underline{90.0}}&37.4 \\
		&Dynamic RCNN \cite{DynamicRCNN}&41.2M&29.4 &44.0 &34.2 &24.4 &{31.2} &45.9 &23.8 &55.0 &37.0 &28.4 &40.9 &34.4 &21.7 &33.8 &47.1 &40.8 &60.0 &35.8 \\
		&SABL \cite{SABL}&41.9M&29.6 &43.3 &35.3 &24.0 &{31.0} &{46.6} &25.5 &77.6 &37.1 &29.3 &42.2 &35.7 &21.8 &34.8 &47.2 &42.6 &{\underline{90.0}} &36.6 \\
		&CenterNet \cite{CenterNet}&14.4M&17.8 &31.7 &18.2 &16.7 &18.4 &29.4 &16.4 &25.0 &28.9 &15.5 &27.3 &16.2 &12.2 &17.7 &25.7 &28.5 &30.0 &28.2 \\
		&FCOS \cite{FCOS}&31.9M&17.5 &28.6 &19.2 &15.3 &18.8 &32.4 &20.0 &45.2 &30.1 &16.9 &27.7 &19.2 &14.0 &17.4 &38.6 &29.0 &{\underline{90.0}} &31.2 \\
		&VarifocalNet \cite{VarifocalNet}&32.5M&26.9 &41.6 &30.1 &22.1 &27.8 &45.0 &27.9 &80.0 &{39.1} &{30.4} &45.8 &36.5 &20.2 &34.7 &{51.4} &{\underline{46.8}} &{\underline{90.0}} &{40.6} \\
		&Deformable DETR \cite{Deformable-DETR}&39.8M&28.2 &{45.4} &32.0 &25.2 &27.2 &43.5 &22.9 &65.0 &38.5 &28.0 &{44.2} &32.7 &21.4 &31.6 &48.1 &37.6 &70.0 &38.8 \\
		&Sparse RCNN \cite{SparseRCNN}&44.2M&19.2 &29.8 &21.9 &18.8 &19.5 &33.4 &14.9 &40.1&33.7 &20.6 &31.5 &24.0 &18.0 &22.4 &31.6 &31.0 &40.0 &36.1 \\
		&{CO-DETR\cite{CO-DETR}}&{65.2M}&{\underline{35.0}}&{52.0}&{40.9}&{\underline{29.1}}&{35.7}&{48.5}&{\underline{34.8}}&{85.0}&{\underline{48.6}}&{\underline{35.0}}&{49.9}&{\underline{42.3}}&{25.0}&{\textbf{40.1}}&{51.2}&{\textbf{48.8}}&{\textbf{100.0}}&{\underline{49.1}}\\
		&{DiffusionDet\cite{DiffusionDet}}&{151.0M}&{\textbf{38.4}}&{\textbf{55.7}}&{\textbf{45.5}}&{\textbf{34.1}}&{\underline{36.4}}&{\textbf{53.5}}&{30.6}&{85.0}&{\textbf{50.4}}&{\textbf{36.9}}&{\textbf{53.4}}&{\textbf{44.1}}&{\textbf{29.5}}&{\underline{39.9}}&{\textbf{57.5}}&{46.7}&{\underline{90.0}}&{\textbf{50.8}}\\
		&{DINO\cite{DINO}}&{97.7M}&{34.7}&{\underline{52.4}}&{\underline{41.3}}&{27.6}&{\textbf{38.6}}&{\underline{52.9}}&{\textbf{45.1}}&{\underline{90.0}}&{\textbf{50.4}}&{34.0}&{50.6}&{41.0}&{25.7}&{37.0}&{53.0}&{44.4}&{80.0}&{48.4}\\
		&{DDQ\cite{DDQ}}&{48.3M}&{31.5}&{47.8}&{37.0}&{26.1}&{31.2}&{48.9}&{34.1}&{85.0}&{47.9}&{34.3}&{49.4}&{41.7}&{\underline{26.4}}&{37.2}&{\underline{55.0}}&{46.0}&{\underline{90.0}}&{50.0}\\
		\hline
		\multirow{3}*{\rotatebox{90}{V-SOD}}&QueryDet \cite{QueryDet}&39.4M&23.6 &36.0 &28.1 &19.2 &25.1 &37.0 &20.3 &30.0 &37.6 &25.1 &39.4 &30.0 &19.3 &31.0 &40.1 &21.3 &40.0 &39.9 \\
		&RFLA \cite{RFLA}&36.3M&{32.1} &{47.1} &{36.3} &{26.8} &30.4 &45.3&{29.4}&50.0&{43.0} &{33.8} &{50.2} &{40.1} &{27.5} &{37.6} &{49.8}&{38.8} &60.0 &{45.4} \\
		&C3Det \cite{C3Det}&55.3M&9.4 &13.8 &11.2 &0.5 &8.1 &42.1 &25.6 &81.7 &10.9 &11.4&16.8&13.3&0.3&8.0&48.7&45.2&{80.0}&13.3\\\hline
		\multirow{3}*{\rotatebox{90}{T-SOD}}&ACM \cite{ACM}&\textbf{{0.4M}}&1.2&2.4&1.0&8.8&5.7&5.0&0.0&0.0&4.9&18.2&32.0&20.5&21.7&25.0&22.1&11.4&40.0&24.3\\
		&ALCNet \cite{ALCNet}&\textbf{{0.4M}}&2.0&5.7&0.5&19.6&9.3&2.0&0.0&0.0&8.6&12.2&22.0&11.9&13.9&15.3&17.8&7.2&19.3&22.0\\
		&DNA-Net \cite{DNAnet}&\underline{{4.7M}}&2.6&4.8&2.5&1.0&4.4&6.2&14.5&0.0&8.3&13.6&22.0&15.1&14.9&13.8&31.3&14.3&0.0&27.7\\
		\hline
		\multirow{8}*{\rotatebox{90}{RGBT-D}}&UA-CMDet (-) \cite{UA-CMDet} &139.2M&18.4&30.2&20.4&11.7&22.5&41.8&21.0&52.1&24.3&24.4&37.9&29.7&14.1&20.9&47.2&42.5&20.3&31.3\\
		&UA-CMDet \cite{UA-CMDet}&139.2M&16.1 &30.6 &15.6 &12.1 &18.5 &32.7 &16.9 &14.2 &26.2 &20.0 &35.0 &21.4 &10.8 &24.1 &35.0 &38.5 &9.7 &28.5 \\
		&ProbEn-early \cite{ProbEn}&60.3M&18.4 &28.3 &22.4 &14.5 &22.0 &32.1 &19.5 &80.0 &23.5 &17.0 &26.8 &19.4 &12.3 &20.2 &32.3 &22.2 &76.7 &22.2 \\
		&ProbEn-middle (-) \cite{ProbEn}&120.6M&24.7 &35.0 &29.9 &20.7 &24.8 &38.1 &26.9 &{85.0} &29.9 &26.0 &36.1 &32.3 &19.2 &31.4 &40.9 &35.8 &{80.0} &31.5 \\
		&ProbEn-middle \cite{ProbEn}&120.6M&27.4 &39.7 &33.4 &23.4 &29.6 &37.2 &{30.8} &{\textbf{92.5}} &33.1 &29.3 &41.0 &35.9 &21.8 &{35.3} &44.1 &37.5 &70.0 &34.0 \\
		&{QFDet \cite{QFDet}}&{60.3M}&{31.8}&{50.0}&{37.5}&{28.1}&{32.6}&{46.4}&{25.2}&{72.5}&{45.3}&{33.2}&{50.3}&{39.3}&{24.3}&{37.5}&{50.9}&{40.6}&{\textbf{100.0}}&{43.0}\\
		&{CALNet \cite{CALNet}}&{153.6M}&{27.5}&{49.4}&{27.1}&{27.1}&{30.5}&{38.8}&{10.3}&{5.0}&{37.3}&{29.4}&{\underline{50.7}}&{30.7}&{25.0}&{35.7}&{44.7}&{24.9}&{5.2}&{38.2}\\
		&{CMA-Det \cite{DVTOD}}&{33.4M}&{18.9}&{40.9}&{16.4}&{19.9}&{19.1}&{27.2}&{12.9}&{54.7}&{35.2}&{18.4}&{36.0}&{16.6}&{17.1}&{21.4}&{23.1}&{24.6}&{70.0}&{31.0}\\
		\hline
	\end{tabular}
	\vspace{-.3cm}
\end{table*}

{Thermal SOD methods can only perform foreground and background segmentation. For performance evaluation on multi-category bboxes, we first enlarge the output channels of CNN-based thermal SOD models to perform multi-category segmentation. Then we employ uniform and Gaussian distribution to generate hard (\textit{i.e.,} all pixels in bboxes are assigned to be positive pixels) and soft (\textit{i.e.,} pixels in bboxes are assigned to probability values under Gaussian distribution \cite{NWD}) mask annotations of each category for training. For test, we transfer the mask-based results to bbox-based results via the minimum enclosing rectangle of each connected region. From the results of Table~\ref{tab-Algorithms}, Soft masks can offer stable training for reasonable evaluation results, and thus breakthrough the evaluation gap between visible and thermal SOD methods.}

Experimental results in Table~\ref{tab-Algorithms} indicate some effective paradigms. 
\textit{1)} End-to-end detection frameworks exhibit significant advancement, including CNN-based detectors \cite{SparseRCNN,SparseRCNN1}, DETR-based detectors \cite{Deformable-DETR,CO-DETR,DINO,DDQ}, Diffusion-based detectors \cite{DiffusionDet}. Due to insufficient appearance cues, dense proposals (e.g., anchor-based \cite{FasterRCNN,CascadeRCNN}, anchor-free \cite{CenterNet,FCOS}, query-based \cite{DDQ,Deformable-DETR}) is more friendly to RGBT-Tiny benchmark. In addition, some technologies (\textit{e.g.,} denoising-based location regression \cite{DINO,DiffusionDet} and contrastive sample balance \cite{DINO,DDQ}) show great potential.
\textit{2)} Small object-specific paradigms demonstrate promising prospects, including region proposal refinement \cite{RFLA,SABL,CascadeRCNN}, multi-scale information fusion \cite{QueryDet,Deformable-DETR,C3Det} and contextual information utilization \cite{QueryDet,C3Det,QFDet}.
\textit{3)} RGBT detection methods can make full use of RGBT complementary information for performance improvements in both modalities, as shown in Table~\ref{tab-Algorithms} and Fig.~\ref{fig_dis_iou} (b). Among them, illumination awareness \cite{UA-CMDet}, misalignment robustness \cite{ProbEn,DVTOD} and semantic-modulation \cite{QFDet,UA-CMDet} can well address the cross-modal semantic conflicts for superior performance. 
\textit{4)} Multi-dimension information (\textit{e.g.,} appearance \cite{CascadeRCNN,Deformable-DETR}, context \cite{QueryDet,C3Det,QFDet}, motion \cite{ProbEn,MOT-QY}) integration benefits the recognition accuracy.

\begin{figure*}[t]
	\vspace{-.15cm}
	\centering
	\includegraphics[width=1\textwidth]{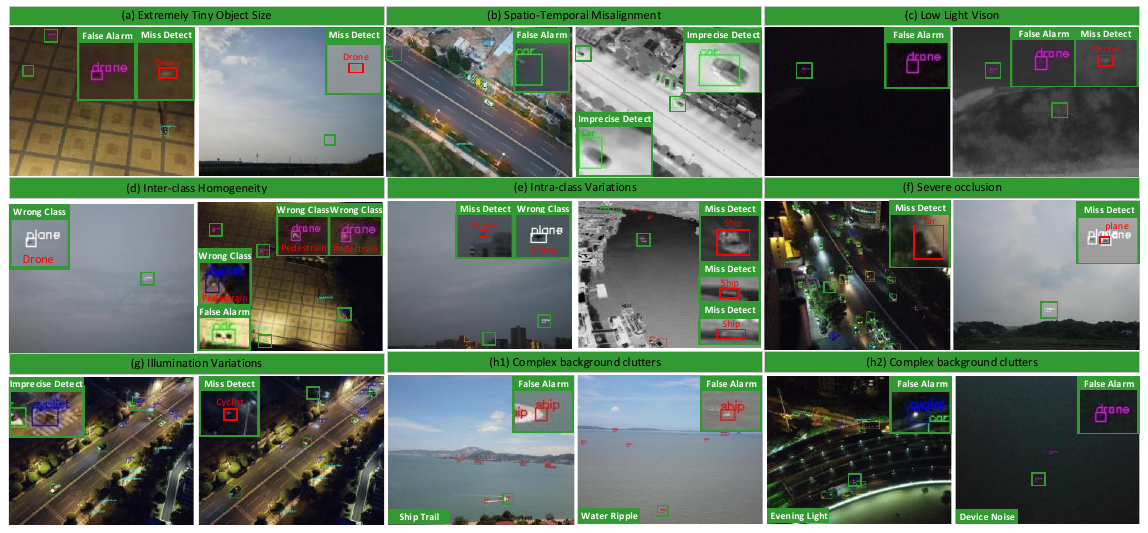}
	\vspace{-.6cm}
	\caption{{Examples of challenging scenes. Green boxes show the zoom-in target regions. Red boxes \& texts inside represent GT bboxes and categories.}
	}\label{fig_challenge} 
	\vspace{-.4cm}
\end{figure*}

{Compared with other public benchmarks \cite{COCO,VisDrone,ACM,DNAnet,ISNet,VTUAV,QFDet}, RGBT-Tiny is an extremely challenging benchmark. 
\textit{1)} Extremely tiny object size with significantly fewer appearance cues raises serious limitations for feature representation learning, and thus results in high miss detection and false alarm, as shown in Fig.~\ref{fig_challenge} (a).
\textit{2)} Cross-modal semantic conflicts, spatio-temporal misalignment and low light vision always lead to severe multimodal fusion errors and significant performance drop, as shown in Fig.~\ref{fig_challenge} (b), (c). 
\textit{3)} Inter-class homogeneity and intra-class variations cause semantic ambiguity for limited recognition performance, as shown in Figs.~\ref{fig_challenge} (d), (e).
\textit{4)} Class imbalance leads to training bias, resulting in limited performance on category with fewer instances (\textit{e.g.,} bus, cyclist), as shown in Fig.~\ref{fig_dis_iou} (c).
Besides the aforementioned challenges, more failure cases primarily stem from severe occlusions (shown in Fig.~\ref{fig_challenge} (f)), illumination variations (shown in Fig.~\ref{fig_challenge} (g)), complex background clutter (shown in Figs.~\ref{fig_challenge} (h1), (h2)). Please refer to supplemental material for detailed discussions of failure cases and potential solutions.} 

\section{Conclusion}
In this paper, we build the first large-scale benchmark dataset (namely RGBT-Tiny) for RGBT-SOD. RGBT-Tiny is an extremely challenging benchmark that contains abundant objects and diverse scenes, and spans over large application scopes, including RGBT image fusion, object detection and tracking. In addition, we propose a scale adaptive fitness measure (SAFit) that exhibits high robustness to both large and small objects, which can provide reasonable performance evaluation and optimal training process. {Based on the proposed RGBT-Tiny dataset, we make comprehensive evaluations on 32 recent state-of-the-art detection algorithms with IoU and SAFit metrics, and summarize the challenges and effective schemes. In future work, we aim to further enlarge the data quantity, refine the annotations, build the foundation model, exploit the temporal information and explore weakly-supervised and unsupervised RGBT SOD.}

{\small
	\bibliographystyle{IEEEtran}
	\bibliography{egbib}
}

\setcounter{section}{0}
\setcounter{figure}{0}
\setcounter{table}{0}

\renewcommand\thesection{\Alph{section}} 
\renewcommand\thetable{\Roman{table}}
\renewcommand\thefigure{\Roman{figure}}



	\twocolumn[{
	\renewcommand\twocolumn[1][]{#1}%
	\begin{center}
		\centering
		\includegraphics[width=1\textwidth]{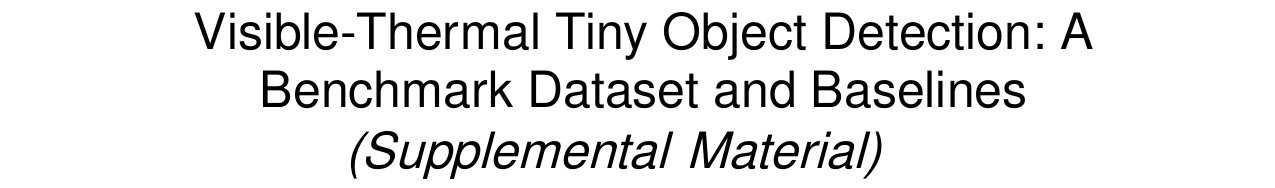}
		\vspace{.05cm}
	\end{center}
	\begin{center}
		\centering
		\includegraphics[width=1\textwidth]{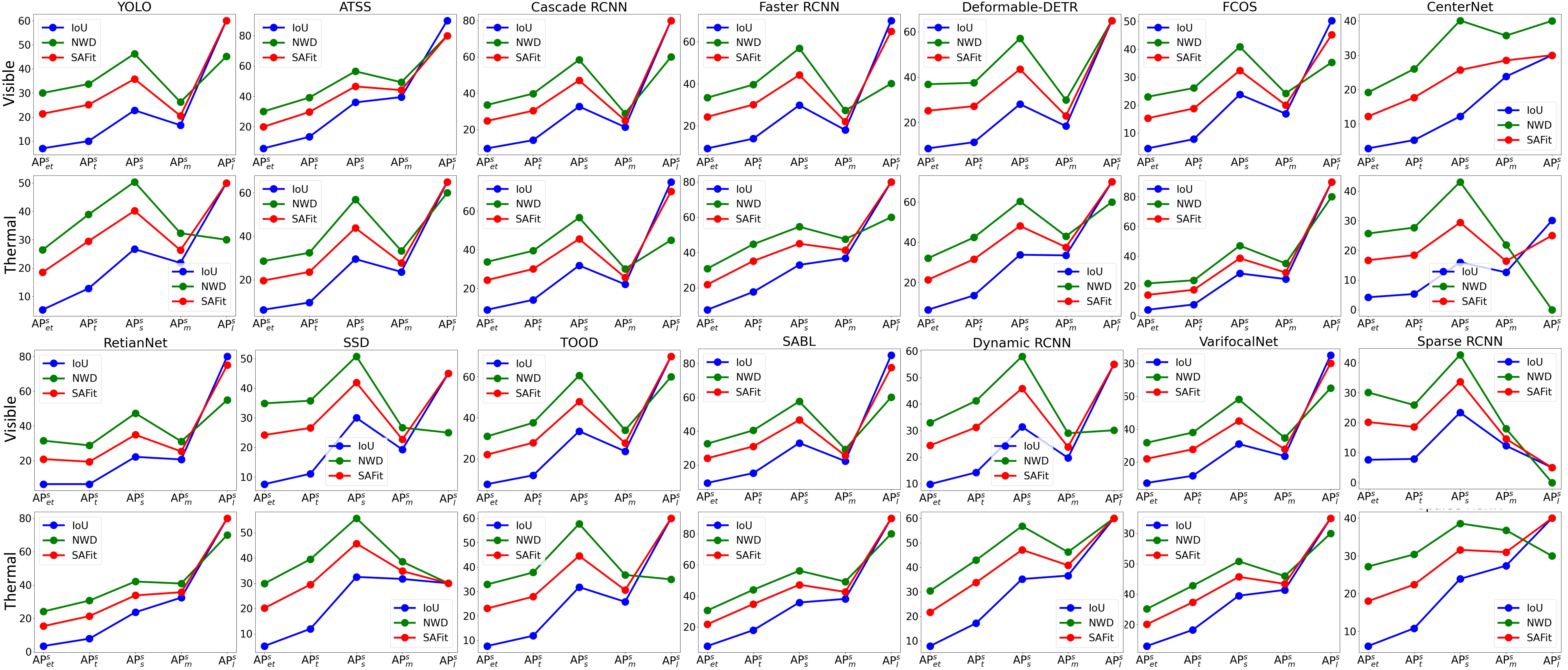}
		\vspace{-0.5cm}
		\captionof{figure}{
			Comparisons among different measures for performance evaluation in visible and thermal modalities under $C$=32. AP$_{et}^s$, AP$_{t}^s$, AP$_s^s$, AP$_m^s$, AP$_l^s$ represent AP values of extremely tiny, tiny, small, medium, large targets.
		}\label{fig_dis_iou_supp}
	\end{center}%
	\vspace{.3cm}
}]

	

	\begin{figure*}[t]
		\centering
		\includegraphics[width=19cm]{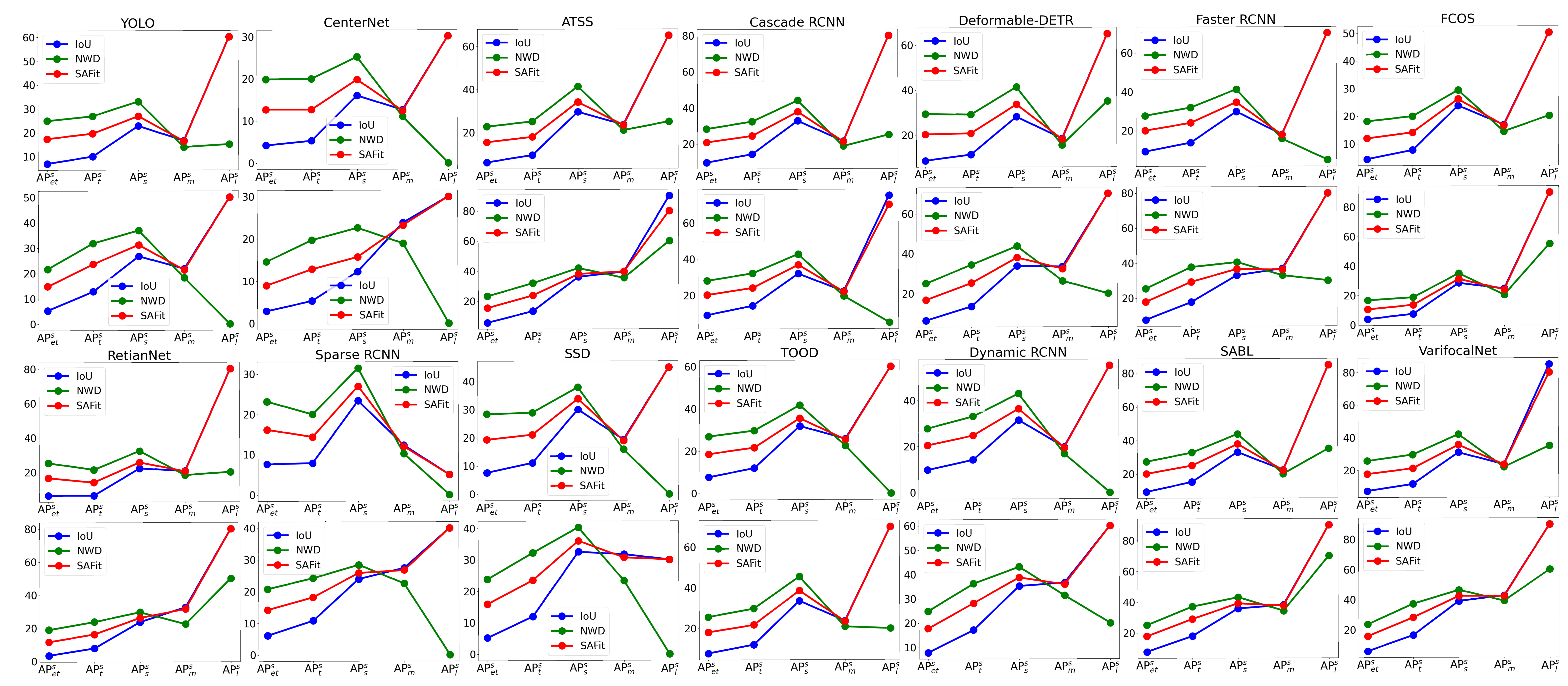}
		\vspace{-.4cm}
		\caption{Comparisons among different measures for performance evaluation in visible and thermal modalities under $C$=16. AP$_{et}^s$, AP$_{t}^s$, AP$_s^s$, AP$_m^s$, AP$_l^s$ represent AP values of extremely tiny, tiny, small, medium, large targets.}\label{fig_dis_iou1_supp} 
	\end{figure*}
	
	\begin{table*}[t]
		\footnotesize
		\centering
		\caption{SAFit-based results of different losses equipped with different detectors. SAFit-s, SAFit$_g$ are used to investigate the effect of direct transition (\textit{i.e.,} loss function is set to NWD when GT box size is smaller than $C$, while set to IoU vice verse) and stronger components (\textit{i.e.,} Sigmoid weighted summation of GIoU and NWD. Best results are shown in boldface, and second best results are underlined.}\label{tab-losses_supp}
		\setlength{\tabcolsep}{2mm}
		\renewcommand\arraystretch{1.1}
		\vspace{-0.2cm}
		\begin{tabular}{|l|ccc|ccccc|c|ccc|ccccc|c|}
			\hline
			\multicolumn{1}{|c}{\multirow{2}*{Loss}}&\multicolumn{9}{|c}{ATSS}&\multicolumn{9}{|c|}{Sparse RCNN}\\\cline{2-19}
			&AP&AP$_{50}$&AP$_{75}$&AP$_{et}^s$&AP$_{t}^s$&AP$_s^s$&AP$_m^s$&AP$_l^s$&AR&AP&AP$_{50}$&AP$_{75}$&AP$_{et}^s$&AP$_{t}^s$&AP$_s^s$&AP$_m^s$&AP$_l^s$&AR\\\hline
			GIoU&24.2 &38.1 &\underline{26.8} &19.6 &23.6 &43.8 &27.5 &65.0&\underline{38.0} &19.2 &29.8 &21.9 &18.8 &\underline{19.5} &33.4 &14.9 &40.1&33.7 \\
			DIoU&23.6 &37.5 &26.1 &19.2 &23.1 &43.6 &\underline{28.2} &\textbf{80.0}&36.8 &19.1 &29.7 &21.7 &19.7 &18.5 &33.9 &14.7 &\underline{60.1}&33.7 \\
			CIoU&24.2 &39.1 &26.0 &19.6 &24.2 &42.6 &\textbf{29.0} &65.0 &37.4&20.0 &30.6 &23.0 &19.5 &19.0 &33.9 &14.8 &\textbf{80.1} &33.9\\
			IoU&23.3 &37.3 &24.9 &19.3 &23.7 &42.3 &26.5 &\underline{75.0}&37.4 &8.1 &14.7 &8.5 &11.3 &7.1 &11.8 &0.2 &0.0&24.1 \\
			NWD&24.1&38.6&26.2&19.6&23.6&\underline{44.2}&27.0&60.0&\textbf{38.4} &19.7 &30.2 &23.2 &19.7 &18.4 &\underline{35.8} &\underline{15.2} &50.0&34.0 \\
			SAFit-s&24.3&39.0&26.2&\underline{19.7}&\underline{24.3}&42.1&27.5&60.0&36.8&19.8&31.2&22.5&18.4&19.4&35.5&14.5&50.1&34.2 \\
			SAFit&\underline{24.5}&\underline{39.2}&26.6&\underline{19.7}&\textbf{25.1}&43.5&28.0&60.1&37.4 &\underline{21.4} &\underline{32.2} &\underline{25.0} &\underline{20.4} &\textbf{20.5} &\textbf{36.3} &14.5 &15.0&\textbf{35.8} \\
			SAFit$_g$&\textbf{24.7}&\textbf{39.3}&\textbf{27.1}&\textbf{20.6}&23.6&\textbf{44.4}&27.5&70.0&37.4&\textbf{22.0} &\textbf{34.0} &\textbf{25.2} &\textbf{22.2} &\underline{19.5}&34.9&\textbf{16.3} &15.0&\underline{34.5} \\
			\hline
		\end{tabular}
	\end{table*}
	
	\begin{table*}[t]
		\footnotesize
		\centering
		\caption{IoU-based results of different losses equipped with different detectors. SAFit-s, SAFit$_g$ are used to investigate the effect of direct transition (\textit{i.e.,} loss function is set to NWD when GT box size is smaller than $C$, while set to IoU vice verse) and stronger components (\textit{i.e.,} Sigmoid weighted summation of GIoU and NWD. Best results are shown in boldface, and second best results are underlined.}\label{tab-losses1_supp}
		\setlength{\tabcolsep}{2mm}
		\renewcommand\arraystretch{1.1}
		\vspace{-0.2cm}
		\begin{tabular}{|l|ccc|ccccc|c|ccc|ccccc|c|}
			\hline
			\multicolumn{1}{|c}{\multirow{2}*{Loss}}&\multicolumn{9}{|c}{ATSS}&\multicolumn{9}{|c|}{Sparse RCNN}\\\cline{2-19}
			&AP&AP$_{50}$&AP$_{75}$&AP$_{et}^s$&AP$_{t}^s$&AP$_s^s$&AP$_m^s$&AP$_l^s$&AR&AP&AP$_{50}$&AP$_{75}$&AP$_{et}^s$&AP$_{t}^s$&AP$_s^s$&AP$_m^s$&AP$_l^s$&AR\\\hline
			GIoU&10.9 &25.3 &8.1 &6.1 &\underline{9.5} &28.6 &22.3 &60.0&\underline{19.1} &9.1 &21.4 &\underline{6.4} &7.3 &\textbf{9.0} &23.3 &\underline{12.6}&45.0&19.2 \\
			DIoU&10.9 &25.6 &8.0 &6.0 &8.9 &28.6 &\underline{24.1} &\textbf{80.0}&18.9 &8.9 &20.9 &6.2 &7.4 &8.1 &23.3 &12.3 &\underline{60.1}&18.9 \\
			CIoU&11.0 &25.9 &\textbf{8.3} &\textbf{6.3} &9.0 &28.3 &\textbf{24.9} &65.0&\underline{19.1} &9.6 &22.7 &\textbf{6.5} &7.8 &\underline{8.8} &23.4&12.3 &\textbf{80.1}&\underline{19.8} \\
			IoU&10.6 &25.0 &8.0 &5.8 &9.2 &28.2 &22.2 &\underline{75.0}&19.0 &2.6 &7.8 &0.9 &3.2 &2.4 &6.5 &0.1 &0.0&11.9 \\
			NWD&10.7&25.8&7.7&5.9&9.0&\underline{29.8}&22.7&60.0&19.2 &8.6 &22.0 &4.6 &7.2 &8.1 &22.8 &12.5 &55.0&18.5 \\
			SAFit-s&10.6&\underline{26.0}&7.5&6.1&8.8&27.9&22.8&60.0&18.2&8.4&21.4&4.9&6.3&8.0&22.6&11.5&50.1&18.6 \\
			SAFit&\underline{11.2}&\textbf{26.9}&\underline{8.1}&\underline{6.2}&\textbf{9.6}&28.6&23.4&60.1&19.0 &\underline{10.0} &\underline{23.9} &\underline{6.4} &\underline{8.3} &\textbf{9.0} &\textbf{25.0} &11.9 &20.0&\textbf{20.4} \\
			SAFit$_g$&\textbf{11.4}&\textbf{26.9}&\textbf{8.3}&\textbf{6.3}&9.3&\textbf{30.2}&23.1&70.0&19.1  &\textbf{10.1} &\textbf{24.4} &\textbf{6.5} &\textbf{8.7} &\underline{8.8} &\underline{23.5} &\textbf{13.3} &15.0&19.5 \\
			\hline
		\end{tabular}
	\end{table*}
	
	\begin{table*}[t]
		\footnotesize
		\centering
		\caption{IoU-based results of existing visible generic detection (V-D), visible SOD (V-SOD), thermal SOD (T-SOD), RGBT detection methods (RGBT-D) methods on RGBT-Tiny dataset. ``\#Param." represents the number of parameters. AP$_{et}^s$, AP$_{t}^s$, AP$_s^s$, AP$_m^s$, AP$_l^s$ represent AP values of extremely tiny, tiny, small, medium, large targets. ``(-)" represents network trained under replicated visible or thermal inputs to investigate the influence of multimodal fusion. Note that, the results of T-SOD are trained with hard (left) and soft (right) masks generated by bboxes under uniform and Gaussian distributions.}\label{tab-Algorithms1_supp}
		\vspace{-.1cm}
		\setlength{\tabcolsep}{1mm}
		\renewcommand\arraystretch{1}
		\begin{tabular}{|c|l|c|ccc|ccccc|c|ccc|ccccc|c|}
			\hline
			&\multicolumn{1}{c|}{\multirow{2}*{Methods}}&\multirow{2}*{\#Param.}&\multicolumn{9}{c|}{Visible}&\multicolumn{9}{c|}{Thermal}\\\cline{4-21}
			&&&AP&AP$_{50}$&AP$_{75}$&AP$_{et}^s$&AP$_{t}^s$&AP$_s^s$&AP$_m^s$&AP$_l^s$&AR&AP&AP$_{50}$&AP$_{75}$&AP$_{et}^s$&AP$_{t}^s$&AP$_s^s$&AP$_m^s$&AP$_l^s$&AR\\\hline
			\multirow{18}*{\rotatebox{90}{V-D}}&SSD \cite{SSD}&25.2M&12.6 &31.0 &8.6 &7.6 &11.1 &30.0 &19.3 &45.0 &19.1 &12.2 &30.4 &7.5 &5.2 &12.0 &32.5 &31.7 &30.0 &18.9 \\
			&YOLO \cite{YOLO}&61.5M&10.0 &26.7 &5.1 &7.0 &10.0 &22.8 &16.6 &60.1 &15.9 &10.6 &27.8 &6.0 &5.3 &12.8 &26.7 &21.8 &50.0 &17.1 \\
			&TOOD \cite{TOOD}&31.8M&13.0 &30.3 &9.9 &7.7 &11.9 &31.8 &25.8 &60.0 &20.2 &13.4 &31.1 &10.3 &7.7 &12.0 &33.5 &23.7 &70.0 &20.7 \\
			&ATSS \cite{ATSS}&31.9M&10.9 &25.3 &8.1 &6.1 &9.5 &28.6 &22.3 &60.0&19.1 &13.8 &32.0 &9.6 &5.7 &13.3 &{36.0} &39.5 &{\underline{90.0}} &21.7 \\
			&RetinaNet \cite{RetinaNet}&36.2M&8.1 &21.7 &4.5 &6.5 &6.5 &22.2 &20.7 &80.1 &16.5 &7.9 &20.3 &5.1 &3.5 &7.9 &23.7 &32.5 &80.0 &17.8 \\
			&Faster RCNN \cite{FasterRCNN}&41.2M&14.1 &32.5 &10.0 &9.3 &13.9 &29.8 &18.0 &70.1 &19.9 &15.4 &35.2 &10.5 &7.6 &17.7 &33.0 &36.8 &80.0 &21.1 \\
			&Cascade RCNN \cite{CascadeRCNN}&68.9M&{15.0} &{33.9} &{11.1} &{9.7} &{14.2} &{32.7} &21.4 &80.0 &{20.5}&14.6 &32.8 &11.2 &{9.1} &14.2 &31.9 &22.2 &75.0 &19.6 \\
			&Dynamic RCNN \cite{DynamicRCNN}&41.2M&14.5 &33.5 &10.7 &{9.8} &{14.2} &31.4 &19.7 &55.0 &20.2 &15.2 &33.7 &11.3 &7.9 &17.2 &35.2 &36.6 &60.0 &20.9 \\
			&SABL \cite{SABL}&41.9M&{15.1} &{33.6} &{11.9} &9.4 &{15.2} &{33.0} &22.3 &{85.0} &20.3 &15.8 &34.6 &{12.4} &7.8 &{18.0} &35.8 &38.1 &{\underline{90.0}} &21.5 \\
			&CenterNet \cite{CenterNet}&14.4M&5.6 &17.3 &2.0 &4.2 &5.3 &16.0 &12.6 &30.1 &11.6 &5.3 &16.1 &1.8 &2.9 &5.3 &12.2 &23.8 &30.0 &12.2 \\
			&FCOS \cite{FCOS}&31.9M&7.6 &18.5 &5.3 &4.5 &7.8 &23.8 &16.9 &50.3 &14.4 &8.2 &19.2 &5.7 &4.0 &7.5 &28.4 &24.6 &{\underline{90.0}} &15.8 \\
			&VarifocalNet \cite{VarifocalNet}&32.5M&13.0 &29.4 &10.1 &7.4 &11.7 &31.0 &23.7 &{85.0} &20.3 &{15.9} &{35.3} &{12.0} &5.9 &16.5 &{39.0} &{42.8} &{\underline{90.0}} &{22.9} \\
			&Deformable DETR \cite{Deformable-DETR}&39.8M&12.3 &30.9 &7.3 &8.6 &11.3 &28.1 &18.4 &65.0 &{20.5} &13.0 &33.1 &7.6 &6.5 &13.6 &33.8 &33.5 &70.0 &21.3 \\
			&Sparse RCNN \cite{SparseRCNN}&44.2M&9.1 &21.4 &6.4 &7.3 &9.0 &23.3 &12.6&45.0&19.2 &10.3 &23.4 &7.8 &6.1 &10.8 &23.9 &27.4 &40.0 &{22.0} \\
			&{CO-DETR \cite{CO-DETR}}&{65.2M}&{\underline{17.8}}&{39.1}&{\underline{14.6}}&{\underline{11.8}}&{17.5}&{34.4}&{\underline{30.6}}&{\underline{90.0}}&{27.2}&{\underline{19.6}}&{\underline{40.9}}&{\textbf{16.6}}&{9.2}&{\textbf{21.9}}&{\underline{40.4}}&{\textbf{45.0}}&{\textbf{100.0}}&{29.8}\\
			&{DiffusionDet \cite{DiffusionDet}}&{151.0M}&{\textbf{19.7}}&{\textbf{42.4}}&{\textbf{16.2}}&{\textbf{13.6}}&{\underline{18.5}}&{\underline{39.9}}&{27.0}&{85.0}&{\underline{28.5}}&{\textbf{19.7}}&{\textbf{41.8}}&{\textbf{16.6}}&{\underline{11.4}}&{\underline{20.4}}&{\textbf{45.5}}&{\underline{42.3}}&{\underline{90.0}}&{29.7}\\
			&{DINO \cite{DINO}}&{97.7M}&{17.0}&{\underline{39.6}}&{11.7}&{9.5}&{\textbf{18.7}}&{4\textbf{0.4}}&{\textbf{40.8}}&{\underline{90.0}}&{\textbf{30.0}}&{17.1}&{39.0}&{12.3}&{8.3}&{18.5}&{39.9}&{40.2}&{\underline{90.0}}&{28.7}\\
			&{DDQ \cite{DDQ}}&{48.3M}&{15.2}&{34.6}&{11.7}&{9.3}&{14.5}&{34.1}&{29.5}&{85.0}&{26.2}&{18.9}&{39.8}&{\underline{15.5}}&{9.6}&{19.5}&{42.6}&{41.8}&{\underline{90.0}}&{\underline{30.6}}\\
			\hline
			\multirow{3}*{\rotatebox{90}{V-SOD}}&QueryDet \cite{QueryDet}&39.4M&10.8 &27.4 &6.5 &5.9 &10.8 &25.5 &16.1 &35.0 &7.6 &11.5 &29.8 &5.8 &4.6 &13.8 &27.9 &17.8 &40.0 &8.3 \\
			&RFLA \cite{RFLA}&36.3M&14.3 &34.0 &9.7 &9.5 &14.0 &31.2 &{25.0} &50.0 &{23.0} &{16.4} &{38.1} &11.4 &8.8 &{18.7} &35.9 &34.4 &60.0 &25.6 \\
			&C3Det \cite{C3Det}&55.3M&6.3 &12.2 &5.7 &0.0 &4.0 &29.2 &21.6 &81.7 &3.7 &7.8 &14.6 &7.8 &0.1 &4.1 &33.8 &{40.8} &80.0 &5.2 \\
			\hline
			\multirow{3}*{\rotatebox{90}{T-SOD}}&ACM \cite{ACM}&\textbf{0.4M}&0.0&0.0&0.0&0.0&0.0&0.0&0.0&0.0&0.0 &6.9 &17.2 &4.2 &3.8 &6.7 &16.6 &23.5 &65.0 &10.2 \\
			&ALCNet \cite{ALCNet}&\textbf{0.4M}&0.0&0.0&0.0&0.0&0.0&0.1&0.0&0.0&0.2 &3.6 &7.5 &3.3 &0.4 &2.8 &8.5 &19.0 &60.0 &4.5 \\
			&DNA-Net \cite{DNAnet}&\underline{4.7M}&0.5&1.5&0.2&0.0&0.9&1.8&9.3&0.0&2.8&2.0 &5.7 &1.0 &1.1 &1.5 &8.7 &3.5 &0.0 &3.8 \\
			\hline
			\multirow{8}*{\rotatebox{90}{RGBT-D}}
			&UA-CMDet (-) \cite{UA-CMDet}&139.2M&8.6 &21.2 &5.5 &3.2 &9.1 &27.5 &17.0 &52.1 &5.7 &11.9 &29.4 &7.1 &3.4 &11.9 &31.6 &38.4 &32.5 &8.5 \\
			&UA-CMDet \cite{UA-CMDet}&139.2M&6.1 &16.7 &3.2 &2.2 &5.7 &19.3 &12.3 &20.1 &5.4 &8.0 &21.0 &4.4 &1.8 &7.5 &18.8 &34.8 &13.1 &5.8 \\
			&ProbEn-early \cite{ProbEn}&60.3M&3.2 &8.7 &1.6 &1.1 &4.3 &5.9 &12.4 &0.0 &0.7 &7.4 &18.9 &4.2 &3.6 &8.3 &21.6 &18.8 &76.7 &5.5 \\
			&ProbEn-middle (-) \cite{ProbEn}&120.6M&12.8 &29.2 &9.1 &8.6 &12.0 &28.5 &{23.9} &{85.0} &8.6 &14.1 &31.3 &10.1 &7.0 &16.4 &30.6 &32.5 &{85.0} &8.5 \\
			&ProbEn-middle \cite{ProbEn}&120.6M&13.3 &32.0 &8.3 &9.0 &13.1 &27.3 &27.5 &{\textbf{92.5}} &19.2 &15.5 &34.6 &11.3 &{8.6} &17.2 &34.0 &34.3 &70.0 &10.1 \\
			&{QFDet \cite{QFDet}}&{60.3M}&{13.6}&{36.0}&{7.7}&{8.8}&{13.6}&{30.6}&{20.6}&{72.5}&{22.1}&{16.6}&{38.9}&{11.2}&{\textbf{13.0}}&{18.1}&{21.3}&{20.0}&{50.0}&{\textbf{31.2}}\\
			&{CALNet \cite{CALNet}}&{153.6M}&{10.6}&{29.2}&{5.0}&{8.1}&{12.6}&{23.0}&{6.8}&{10.0}&{16.9}&{12.7}&{33.1}&{7.0}&{8.0}&{15.7}&{28.4}&{20.1}&{5.4}&{18.6}\\
			&{CMA-Det \cite{DVTOD}}&{33.4M}&{5.5}&{16.1}&{2.2}&{3.3}&{5.7}&{15.2}&{9.1}&{54.7}&{13.0}&{5.6}&{16.9}&{2.5}&{3.1}&{6.2}&{8.0}&{21.5}&{70.0}&{12.6}\\
			\hline
		\end{tabular}
	\end{table*}
	
	\noindent
	First, we provide more evaluation results of SAFit measure for the analyses in Section 4.1-\textit{SAFit Measure for Evaluation}. Second, we provide more SAFit-based and IoU-based results of SAFit loss in RGBT-Tiny datasets for the analyses in Section 4.1-\textit{SAFit Loss for Training}. {Fourth}, we provide IoU-based results of RGBT-Tiny benchmark for the analyses in Section 4.2. {Finally, we make discussions on some typical failure cases of existing state-of-the-art algorithms and provide some potential solutions.}
	
	\section{Results of SAFit Measure for Evaluation}\label{sec:SAFit-evaluation}
	We employ IoU, NWD and SAFit measures under different $C$ values (\textit{i.e.,} $C$=$16$, $C$=$32$) for performance evaluation on 14 recent state-of-the-art visible generic detection methods \cite{SSD,YOLO,TOOD,ATSS,RetinaNet,FasterRCNN,CascadeRCNN,DynamicRCNN,SABL,CenterNet,FCOS,VarifocalNet,Deformable-DETR,SparseRCNN}. AP results in visible and thermal modalities across different target scales (\textit{i.e.,} extremely tiny, tiny, small, medium, large) are shown in Figs.~\ref{fig_dis_iou_supp} and \ref{fig_dis_iou1_supp}. It can be observed that AP values of SAFit are closer to those of NWD when GT bbox size is smaller, and approximate those of IoU as size increases. In addition,  $C$ controls the maximum values of NWD and SAFit, and a higher $C$ value results in higher NWD and SAFit values. Moreover, $C$ controls the transition point of dominated measures. That is, when GT bbox size is smaller than $C$, NWD occupies more proportions of SAFit, while a larger GT bbox size results in IoU domination. In conclusion, extensive experimental results have demonstrated that SAFit can make comprehensively reasonable evaluations on both large and small targets, and can flexibly control the transition point by a size-aware parameter $C$ for custom requirements.
	
	\section{Results of SAFit Loss for Training}\label{sec:SAFit-loss}
	Tables~\ref{tab-losses_supp} and~\ref{tab-losses1_supp} provide SAFit-based and IoU-based results achieved by different detectors (\textit{i.e.,} ATSS \cite{ATSS}, Sparse RCNN \cite{SparseRCNN}) with different losses (\textit{i.e.,} IoU \cite{Fpillars}, DIoU \cite{DIoU}, CIoU \cite{DIoU}, GIoU \cite{GIoU}, NWD \cite{NWD} and SAFit losses) in visible modality of RGBT-Tiny dataset. Note that, we employ 2 variants of SAFit loss (\textit{i.e.,} SAFit-s and SAFit$_g$) to investigate the performance of direct transition (\textit{i.e.,} loss function is set to NWD when GT box size is smaller than $C$, while set to IoU vice verse) and stronger component (\textit{i.e.,} Sigmoid weighted combination of GIoU and NWD). It can be observed that SAFit loss shows high robustness to different detectors under both IoU and SAFit metrics. In addition, compared with SAFit-s loss, SAFit loss achieves higher AP values, which demonstrates that size-aware weighted summation is superior to direct transition. Moreover, SAFit$_g$ performs superior to SAFit, which demonstrates that stronger components further promote the detection performance. Furthermore, compared with its components, SAFit$_g$ and SAFit achieves higher AP$_{et}^s$, AP$_{t}^s$ values than NWD, and higher AP$_m^s$ values than GIoU and IoU respectively, which demonstrates that our size-aware weighted summation can not only combine both advantages of its components but also produce a breakthrough by offering a more stable and smooth training for targets with varied sizes.
	\vspace{-.2cm}
	
	\section{IoU-based Results of RGBT-Tiny Benchmark}\label{sec:IoU-results}
	Table~\ref{tab-Algorithms1_supp} provides IoU-based results of our benchmark. It can be observed that IoU-based results are lower than SAFit-based results, and the differences result from unreasonable performance evaluation of IoU-based metrics on small objects.
	
	\section{{Typical Failure Cases and Potential Solutions}}\label{sec:CS&FD}
	\begin{figure*}[t]
		\centering
		\includegraphics[width=1\textwidth]{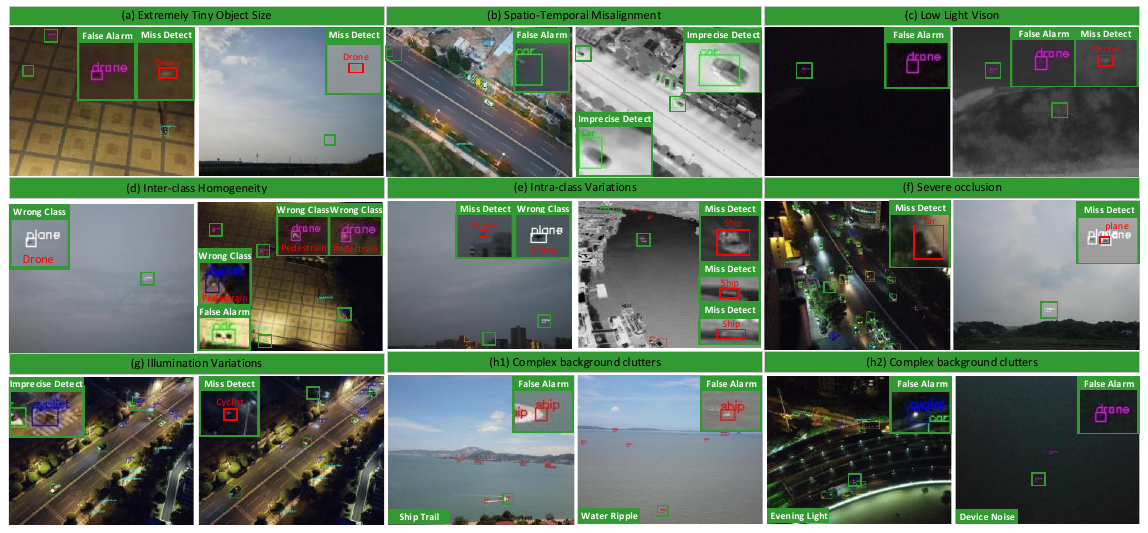}
		\vspace{-.5cm}
		\caption{{Failure cases of existing state-of-the-art algorithms. Green boxes show the zoom-in target regions. Red boxes \& texts inside represent GT boxes and classes.}}\label{fig_failure}
	\end{figure*}
	
	{Figure~\ref{fig_failure} visualizes some typical failure cases of existing state-of-the-art algorithms. Detailed discussions are as follows with some potential solutions being provided.}
	
	\begin{enumerate}
		\item {\textbf{Extremely tiny object size:} As shown in Fig.~\ref{fig_failure} (a), the extremely small object size with significantly fewer appearance cues raises serious limitations for feature representation learning, and thus results in high miss detection and false alarm. This limitation highlights the need for exploring multi-dimensional reliable information (\textit{e.g.,} appearance, context, motion, radiation information) for object detection.}
		
		\item {\textbf{Spatio-Temporal Misalignment:} As shown in Fig.~\ref{fig_failure} (b), the inherent disparity variation of dual lenses \cite{disparities} presents significant challenges in RGBT fusion, and thus leads to suboptimal detection performance. This limitation highlights the need for dynamic cross-modal feature alignment.}
		
		\item {\textbf{Low Light Vision:} As shown in Fig.~\ref{fig_failure} (c), extreme asymmetric visible and thermal information of low light vision (\textit{e.g.,} low and invisible light vision) introduce difficulties in RGBT fusion, resulting in detection performance degradation. This limitation highlights the need for illumination-aware cross-modal feature compensation.}
		
		\item {\textbf{Inter-class Homogeneity:} As shown in Fig.~\ref{fig_failure} (d), similar feature representation across classes results in semantic ambiguity, over-fitting in training data and ambiguous decision logic, and thus leads to limited recognition performance, poor interpretability and inferior generalization. This limitation highlights the need to explore more discriminative features (\textit{e.g.,} appearance, context, motion, radiation information and multi-dimensional combination) for object recognition.}
		
		\item {\textbf{Intra-class Variations:} As shown in Fig.~\ref{fig_failure} (e), object appearance can be significantly changed due to complex object states (\textit{e.g.,} shape, texture, color and posture) and observation states (distance, orientation, light, weather, sensor motion, sensor noise), resulting in miss detection and false alarms. This limitation highlights the need for long-term temporal information exploration to assist object detection and recognition.}
		
		\item {\textbf{Severe occlusion:} As shown in Fig.~\ref{fig_failure} (f), algorithms often struggle to accurately identify and localize the object of interest when partially or completely occluded, resulting in a significant drop in detection rates. This limitation highlights the need for occlusion context reasoning, including leveraging generative models to infer the shape and appearance of occluded objects, and incorporating physical and geometric constraints to predict the likelihood of occlusion. In addition, the interaction or occlusion of densely arranged objects often leads to performance drop. This limitation highlights the need for effective disentangling techniques (such as image super-resolution techniques to integrate spatial prior knowledge and graph-based representations for object reconstruction), and temporal consecutive reasoning.}
		
		\item {\textbf{Illumination Variations:} As shown in Fig.~\ref{fig_failure} (g), Illumination variations, particularly in night-time scenarios, pose significant challenges in object detection and recognition systems. This limitation highlights the need for image enhancement and multimodal fusion to improve the robustness.}
		
		\item {\textbf{Complex background clutters:} As shown in Figs.~\ref{fig_failure} (h1), (h2), complex background clutters (\textit{e.g.,} ship trail, water ripple, evening lights and device noise) significantly affect the detection of small objects, resulting in high false alarms. This limitation highlights the need for more robust and adaptive algorithms that can effectively cope with the complexities and diversities of real-world imagery.}
	\end{enumerate}

{\small
	\bibliographystyle{IEEEtran}
	\bibliography{egbib}
}

\end{document}